\documentclass[10pt,twocolumn,letterpaper]{article}

\usepackage{iccv}
\usepackage{times}
\usepackage{epsfig}

\usepackage[accsupp]{axessibility}

\usepackage{graphicx}
\usepackage{rotating}
\usepackage{amsmath}

\usepackage{amssymb}
\usepackage{booktabs}
\usepackage{multirow}
\usepackage{subcaption}
\graphicspath{ {images/} }
\usepackage{blindtext}
\usepackage{multicol}
\usepackage{pifont}
\newcommand{\cmark}{\ding{51}}%
\newcommand{\xmark}{\ding{55}}%

\usepackage[dvipsnames]{xcolor}

\usepackage[breaklinks=true,bookmarks=false]{hyperref}

\iccvfinalcopy 

\def\iccvPaperID{****} 

\ificcvfinal\fi

\begin{document}

\title{Localizing Moments in Long Video Via Multimodal Guidance}

{\author{
Wayner Barrios$^{1}$,
\quad Mattia Soldan$^{2}$, 
\quad Alberto Mario Ceballos-Arroyo$^{3}$,\\
\quad Fabian Caba Heilbron$^{4}$,
\quad Bernard Ghanem$^{2}$ \\\\
\small$^{1}$ Dartmouth \quad $^{2}$King Abdullah University of Science and Technology (KAUST)
\small$^{3}$Northeastern University \quad $^{4}$Adobe Research
}}
\maketitle
\ificcvfinal\thispagestyle{empty}\fi

\begin{abstract}

The recent introduction of the large-scale, long-form MAD and Ego4D datasets has enabled researchers to investigate the performance
of current state-of-the-art methods for video grounding in the long-form setup, with interesting findings: current grounding methods alone fail at tackling this challenging task and setup due to their inability to process long video sequences. In this paper, we propose a method for improving the performance of natural language grounding in long videos by identifying and pruning out non-describable windows. We design a guided grounding framework consisting of a Guidance Model and a base grounding model. The Guidance Model emphasizes describable windows, while the base grounding model analyzes short temporal windows to determine which segments accurately match a given language query. We offer two designs for the Guidance Model: Query-Agnostic and Query-Dependent, which balance efficiency and accuracy. Experiments demonstrate that our proposed method outperforms state-of-the-art models by 4.1\% in MAD and 4.52\% in Ego4D (NLQ), respectively. Code, data and MAD's audio features necessary to reproduce our experiments are available at: \href{https://github.com/waybarrios/guidance-based-video-grounding}{https://github.com/waybarrios/guidance-based-video-grounding}


    
\end{abstract}
\section{Introduction}
\label{sec:intro}
\begin{figure}[!t]
    \centering
    \includegraphics[width=1.0\linewidth]{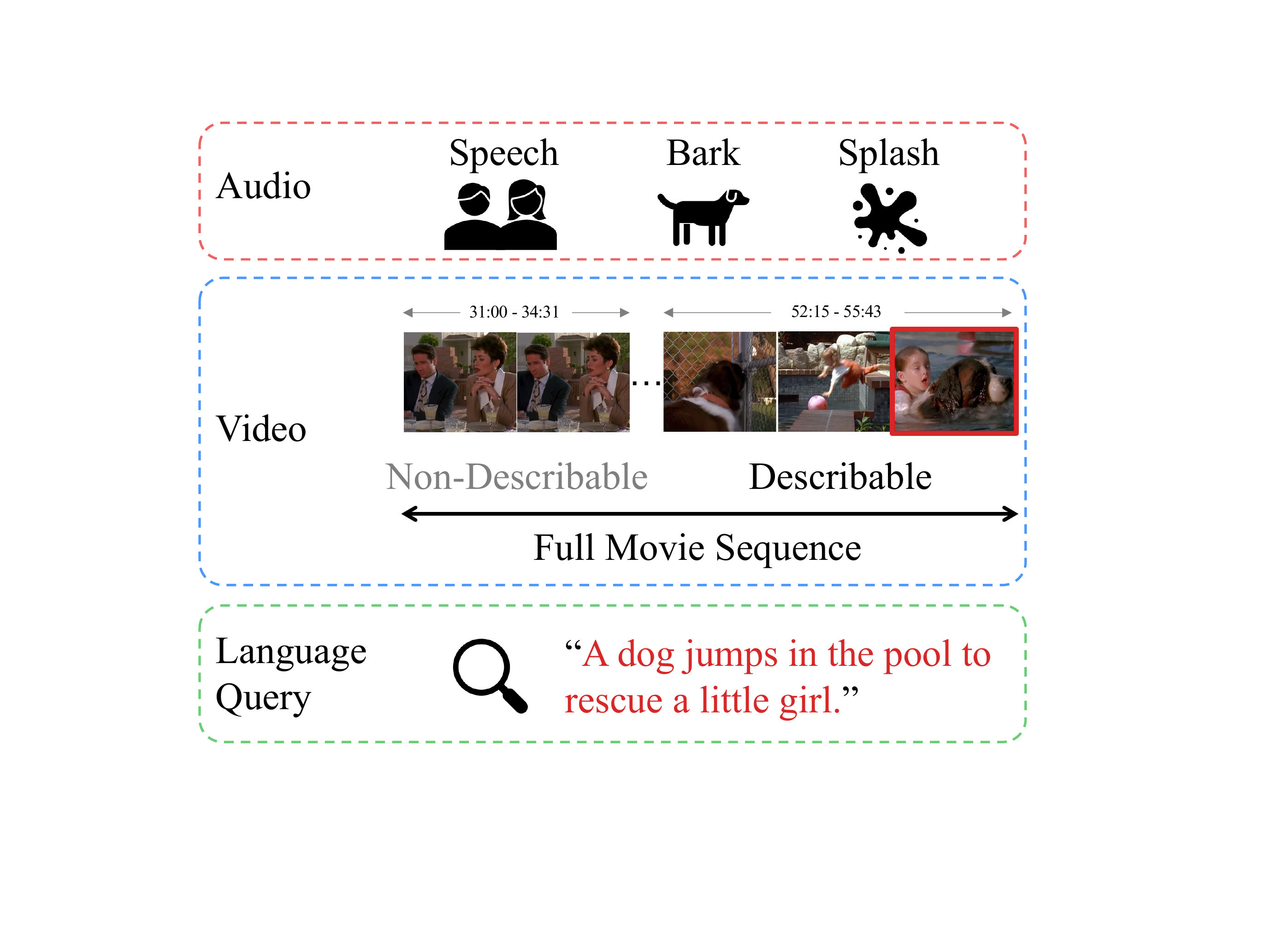}
    \caption{\small\textbf{Describable Window.} We depict the difference between describable and non-describable windows. The former are temporal windows with relevant visual and auditory events that are likely to contain one or more noteworthy moments. The latter can be categorized as ``boring'' video segments (temporal windows) where little happens and no moment of interest happens for grounding.}
    \label{fig:teaser}
\end{figure}

The task of grounding natural language in long-form videos  within large-scale datasets, such as MAD~\cite{Soldan_2022_CVPR} and Ego4D~\cite{ego4d}, can be particularly challenging due to the possibility of encountering many video segments that do not contain any interesting or relevant moments to search for. Given the large search space, it is critical to develop approaches that can quickly scan the video and identify query-able moments via natural language.

For instance, Figure~\ref{fig:teaser} illustrates various moments occurring in a long-form video, where certain moments could have greater relevance than others based on the video's storytelling. The dialogue moment, which spans more than three minutes, would only be described as~\textit{``two people sitting at a table"}, whereas the action scene contains many moments that can be described, such as~\textit{``a girl falling into a pool"},~\textit{``a dog opening a fence"}, and \textit{``a dog jumps to rescue a little girl"}. This demonstrates that certain slices within a video can contain numerous notable moments that users would be highly interested in searching for.

In consequence, we introduce the concept of \textit{Describable windows} (as shown in Figure \ref{fig:teaser}), which are video slices shorter than the original long-form video and have a high probability of containing remarkable and relevant visual moments that can be queried through natural language. Conversely, video slices that do not contain such visually-rich moments that cannot be queried through natural language are considered \textit{non-describable windows}.\\


Our work builds on the hypothesis that certain video segments (\textit{non-describable windows}) are difficult to narrate effectively by natural language, which can lead to noisy predictions when retrieving moments via natural language queries, particularly for long videos. In fact, state-of-the-art grounding models achieve remarkable performance when analyzing short videos~\cite{TACoS_ACL_2013, Krishna_2017_ICCV, Mun_2020_CVPR}; however, their capabilities seem to saturate when tested on longer videos~\cite{Soldan_2022_CVPR}. For example, VLG-Net~\cite{soldan2021vlg} achieves only marginal gains compared to a simple yet effective zero-shot approach based on CLIP~\cite{radford2021learning}, and some models outright fail to achieve reasonable performance\footnote{We trained Moment-DETR~\cite{moment-detr} from scratch on the MAD~\cite{Soldan_2022_CVPR} dataset and found that, even though it achieves good grounding performance in short videos, it utterly fails in long videos. More details can be found in the supplementary material.}.   The above statements highlight the need to detect video slices corresponding to the ``important" parts of a long-form video, \textit{i.e.}, those that contain as many moments as possible. Our intuition is that when analyzing the entire long-form video,  grounding models might overlook the most critical segments of the video, which represents a gap in the current research and motivates us to develop  more sophisticated techniques that can capture certain intricacies using language cues or audio cues on top of the visual information. 

To address this issue, we propose a guided grounding framework~\cite{Hendricks_2017_ICCV, escorcia2019temporal} comprised of two core components: a \textit{Guidance Model} that specializes emphasizing describable windows and a base grounding model that analyzes short temporal windows to determine which temporal segments match a given query accurately. 

Observing multimodal cues is key for detecting describable windows. For instance, let us suppose we want to find the moment when~\textit{``the dog jumps to rescue the little girl''} (Figure~\ref{fig:teaser}). Here, the water splash (sound) and dog jumping (visual) are hints suggesting that lots of visual activities are happening in the scene. This intuition motivates the multimodal design of our Guidance Model. In practice, our model jointly encodes video and audio over an extended temporal window using a transformer encoder~\cite{NIPS2017_attent}.

The proposed Guidance Model can also be used in two different frameworks: \textit{Query Agnostic} and \textit{Query Dependent}. The former pre-computes which parts of a video have a low-probability of containing describable windows, making it suitable for real-time applications or limited computational resources. In contrast, the latter provides more precise results by identifying irrelevant parts of a video based on a given text query, at the cost of being more computationally expensive as the number of queries grows. We also highlight the fact that the two-stage approach can be adapted to any grounding method, making it a flexible and versatile option for a wide range of applications. \\

In summary, our \textbf{contributions} are three-fold: 
\begin{enumerate}
    \item We propose a two-stage guided grounding framework: a general approach for boosting the performance of current state-of-the-art grounding models in the long-form video setting.
    \item We introduce a Guidance Model that is capable of detecting describable windows: a temporal window that contains one or more noteworthy moments.
    \item Through extensive experiments, we empirically show the effectiveness of our two-stage approach. In particular, we validate the benefits of leveraging a Guidance Model that specializes in finding describable windows. We improve state-of-the-art performance on the MAD dataset~\cite{Soldan_2022_CVPR} by $4.1\%$ and improve the performance of an Ego4D baseline model~\cite{ego4d,vslnet} in the NLQ task by $4.52\%$.
\end{enumerate}

\section{Related Work}
\label{sec:related_work}

\noindent\textbf{Video Grounding Methods.}
Video grounding methods can be divided into two families of approaches: \textbf{(i)} proposal-based~\cite{Gao_2017_ICCV,Hendricks_2017_ICCV,soldan2021vlg, 2DTAN_2020_AAAI, xu2023boundary} and \textbf{(ii)} proposal-free~\cite{moment-detr, Zeng_2020_CVPR, Mun_2020_CVPR, chenhierarchical, Rodriguez_2020_WACV, Li_Guo_Wang_2021}. Proposal-based methods rely on producing confidence scores or alignment scores for a previously generated set of $M$ candidate temporal moments $\{(\tau_{start},\tau_{end})\}^M_1$.
On the other hand, proposal-free methods aim at directly regressing the temporal interval boundaries for a given video-query pair. For instance, Mun \etal~\cite{Mun_2020_CVPR} tackle this problem by using a temporal attention-based mechanism that exploits local and global information in bimodal interactions between video segments and semantic phrases in the query to regress the target interval. Li~\etal~\cite{Li_Guo_Wang_2021} uses a pyramid network architecture to leverage multi-scale temporal correlation maps of query-enhanced video features as the input to a temporal-attentive regression module. 

Regarding efficiency, proposal-free approaches have faster inference time since they do not require exhaustively matching the query to a large set of proposals. Conversely, proposal-based methods often provide higher performance at the cost of a much slower inference time. In fact, these methods produce predictions for a large number of temporal proposals and successively apply expensive Non-Maximum Suppression (NMS) algorithms to improve the ranking. 

Soldan~\etal~\cite{Soldan_2022_CVPR} showcased how current state-of-the-art grounding methods fail to tackle the long-form grounding setup where videos are up to several hours in duration. Our focus is thus to design a pipeline able to boost these methods through a two-stage cascade approach. In the experimental section, we will showcase how our flexible design is able to boost both proposal-based and proposal-free methods to achieve a new state-of-the-art performance. \\

\noindent\textbf{Long-form Video Grounding.}
In long-form video understanding, grounding approaches are limited by the inability to process the entire video at once. Such limitation stems from the large computational budget for deep learning-based video perception algorithms and the limited computational resources currently available. Recently, \cite{Soldan_2022_CVPR} proposed to process the video in short windows to enable grounding in long-form videos. Long videos are therefore sliced in overlapping temporal windows, and grounding methods are used to generate predictions within each window. Finally, predictions across all windows are gathered and ranked according to the confidence scores. 

While video slicing is a potential solution for long-form video grounding, it has a significant limitation as grounding methods may introduce a considerable number of false positives in the predictions for any long-form videos. We address this challenge by introducing a two-stage approach that guides any grounding model in making accurate predictions and reducing the presence of false positives. \\

\noindent\textbf{Multimodal Transformers.}
The great flexibility of the transformer architecture~\cite{NIPS2017_7181} has promoted its adoption in several different fields: computer vision~\cite{dosovitskiy2020image,li2022efficientformer}, natural language processing~\cite{bert, brown2020language, wolf2020transformers}, and audio processing~\cite{verma2021audio, niranjan2020end}.
Concurrently, several works have exploited this architecture to process multimodal data~\cite{luo2021clip4clip, radford2021learning, Lei2021LessIM, Li2020HeroHE, xu2021videoclip, carion2020end, Kamath2021MDETRM, ramazanova2023owl}. The key to successfully applying transformers to multimodal data is to learn a projection to a shared embedding space for each element of each modality (frames, spectrograms, language tokens). Our design for the Guidance model, therefore, follows these recent advancements. In particular, we aim to exploit the flexibility of transformers to design a unified architecture able to gather relevant cues from different modalities. We detail the design choices for our Guidance model in the following section. \\

\noindent\textbf{Temporal Proposals.}
Temporal proposals refer to the identification and localization of specific actions or events in an untrimmed video by identifying temporal segments that are likely to contain them.~\cite{SSN2017ICCV,SST}  generate a single video snippet per each specific event followed by an action classifier, which makes it dependent on the label class. The existence of several video snippets that correspond to specific events in a long-form video, where multiple events can occur simultaneously, significantly adds to the complexity of the task. For a given time segment $t$, $N$ video snippets would be required, where $N$ represents the number of concurrent events. 

Recently,~\cite{AAAIYangWW0XYH22,HAN2022217,qing2021temporal} leverages  mechanisms that capture semantic information of video clips, self-attention, as well as local and global temporal relationships using visual information. Although these methods have yielded outstanding outcomes, the nature of grounding in long-form videos necessitates an examination of supplementary cues from other modalities to identify as many events as possible that can be referenced in a shorter video duration, i.e., in a describable window.


\section{Method}
\label{sec:method}
\begin{figure}[!t]
\centering
  \centering
  \includegraphics[width=\linewidth]{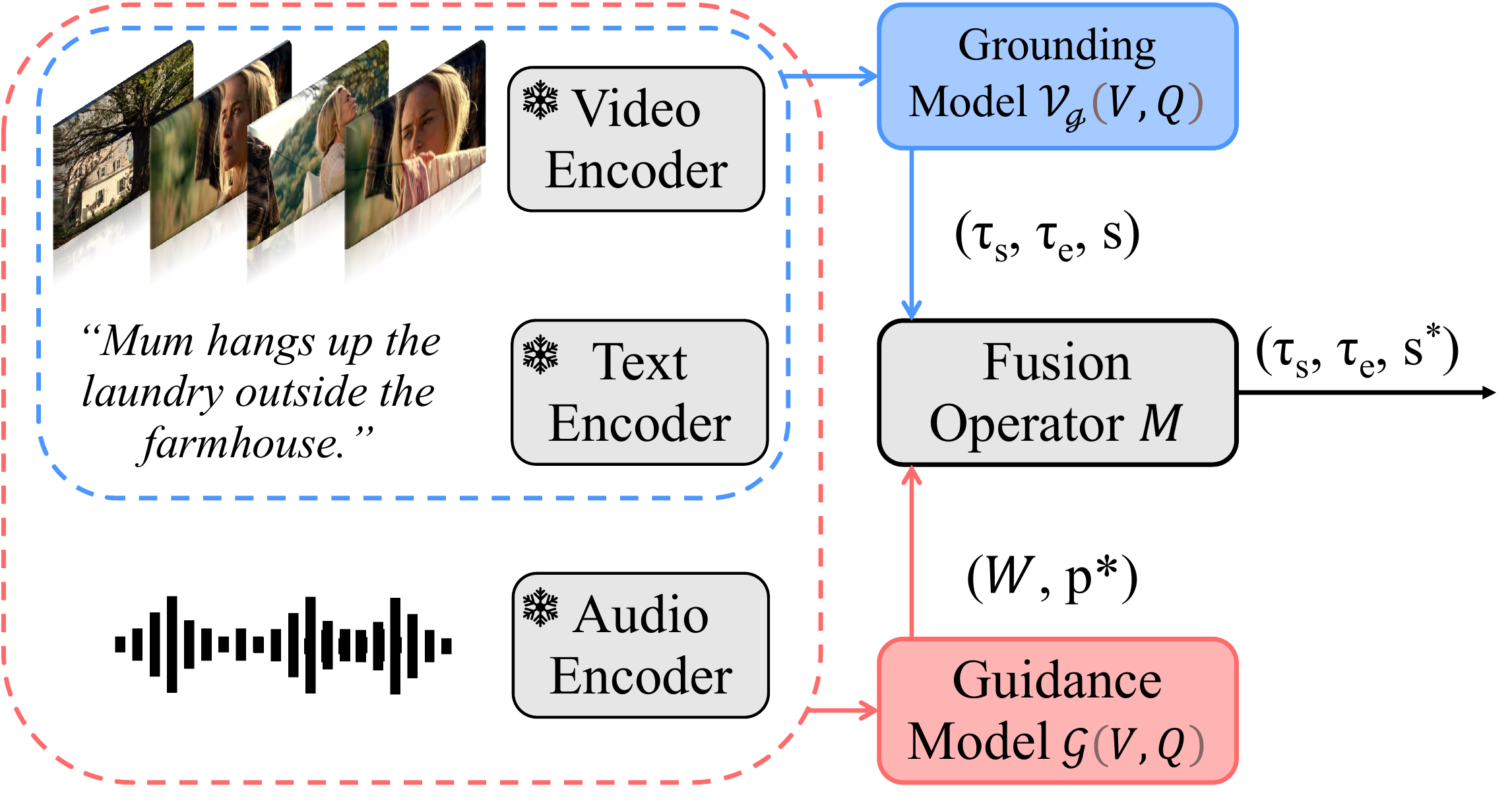}
   \caption{\small\textbf{Guided Grounding pipeline.} Two-stage approach comprised of a video-language grounding model and a guidance model. Given a set of predictions $\{(\tau_{s},\tau_{e}, s)\}^M_1$ from a grounding model, and  
   plausibility score $p^*$, we generate a refined set of predictions  $\{(\tau_{s},\tau_{e}, s^*)\}^M_1$.
   }
   \label{fig:arch}
\end{figure}
Our goal is to train a guidance model $\mathcal{G}$ that finds describable windows that might contain a moment to be grounded in a long video. By finding these windows, we aim to guide  an existing grounding model $\mathcal{V}_g$ in improving its predictions. Let $\mathcal{V}_g$'s temporal moments predictions be defined as: $\{(\tau_{s},\tau_{e}, s)\}^M_1$, where  $(\tau_{s},\tau_{e})$ corresponds to start/end time and $s$ is the noisy confidence score; given a temporal window $W$ sampled from a long video, and the fusion operator $\mathcal{M}$, we seek a model that generates temporal moment predictions such that:

\begin{equation}
    \mathcal{M}\left(\mathcal{V}_g, \mathcal{G}, W\right)\rightarrow\{(\tau_{s},\tau_{e}, s^*)\}^M_1,
\end{equation}

where $s^*$ denotes a refined confidence score for the temporal moment. \\

\noindent \textbf{Grounding model $\mathcal{V}_g$.} The grounding model takes as input a video observation $V$ sampled from a temporal window $W$, and a natural language query $Q$; it then predicts $M$ temporal moments such that:

\begin{equation}
    \mathcal{V}_g\left(V,Q\right)\rightarrow\{(\tau_{s},\tau_{e}, s)\}^M_1.
\end{equation}

In this work, we leverage several pre-trained grounding models such as \cite{moment-detr,Soldan_2022_CVPR,soldan2021vlg,vslnet}.
While these works achieve remarkable performance on localizing \textit{existing} moments within short temporal window, they suffer severely from false positive predictions on windows that do not contain any moment. As we will show in our experiments, modulating the confidence scores of a grounding model can greatly improve the overall performance of the model.\\

\noindent \textbf{Guidance model $\mathcal{G}$.} We seek to train a guidance model that gives low scores to predictions made on empty windows (w/o moments) and gives higher scores to predictions derived from windows containing one or more moments (\textit{describable windows}). Similar to the grounding model, $\mathcal{G}$ ingests a video observation $V$ sampled from $W$, and the target query $Q$. For a given window, we obtain a plausibility confidence score $p^*$ as follows:
\begin{equation}
    \mathcal{G}\left(V,Q\right)\rightarrow p^*.
\end{equation}
\\
\noindent \textbf{Guided grounding output $\mathcal{M}$.} After obtaining a set of predictions $\{(\tau_{s},\tau_{e}, s)\}^M_1$, and a plausibility score $p^*$ for window $W$\!, the fusion operator $M$ aims to generate a refined confidence score for each predicted moment (Figure \ref{fig:arch}). To do so, we simply multiply the plausibility score $p^*$ with each confidence score $\{s\}^M_1$. This generates the final (and refined) set of predictions:

\begin{equation}
    P=\{(\tau_{s},\tau_{e}, s^*)\}^M_1
\end{equation}
\\
\noindent \textbf{Grounding in long-form videos.}
We follow \cite{Soldan_2022_CVPR} and generate predictions in long-form videos by sliding a short window throughout time. Assuming we analyze $K$ windows, our model generates $K{\times}M$ predictions for a natural language query $Q$. We sort the resulting predictions by their confidence scores $s^*$\!.\\

\subsection{Guidance Model Design and Training Details}
\begin{figure}[t]
  \centering
  \includegraphics[width=0.47\textwidth]{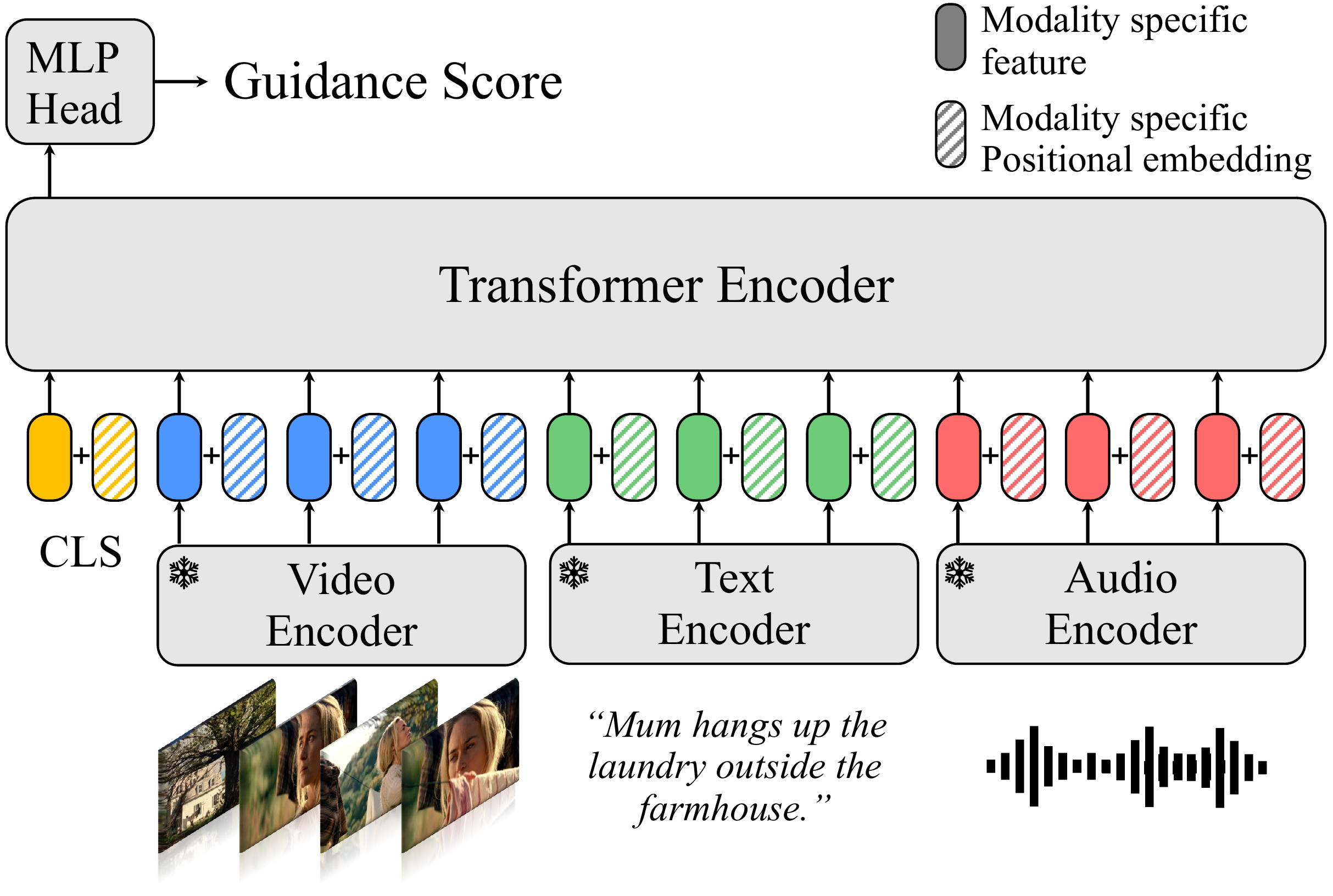}
   \caption{\small\textbf{Guidance model.} Our guidance model can process different modalities (setup-dependent). Each modality representation is provided with a modality-specific positional embedding before being fed to the Transformer Encoder module. The output of the CLS token is projected to a real value, used as a guidance score to condition the grounding models. }
   \label{fig:filter}
\end{figure}
The Guidance Model is depicted in Figure \ref{fig:filter}. Our goal is to train the model $\mathcal{G}$ on a dataset $\mathcal{D}$ containing binary labels (of window plausibility) for a set of temporal windows and their corresponding natural language queries. \\

\noindent\textbf{Window Representation.}
We sample an observation $V$ from an input window $W$. Such observation includes diverse inputs (\textit{i.e.}, video, audio, and text). The use of each modality is setup-dependent and will be thoroughly ablated in Section~\ref{ablation:describabe}.  In practice, we employ frozen encoders to extract embeddings for each modality. Formally, we denote video embeddings, audio embeddings, and textual embeddings as 
$E_v \in \mathbb{R}^{L_{vg} \times D_{v}}$, 
$E_a \in \mathbb{R}^{L_{ag} \times D_{a}}$, and 
$E_t \in \mathbb{R}^{L_{tg} \times D_{t}}$, respectively, where $D_{v}$, $D_{a}$, and $D_{t}$ represent the dimensionality of the features.
Input features are projected to a shared embedding space of size $d_g$ by an MLP projection followed by a layer normalization module~\cite{layernor}. Dropout is used for regularization. 

Our design allows our Guidance Model to be query-dependent or query-agnostic. The inputs for each of these variants are as follows:
\begin{itemize}
    \item[] For \textbf{query-dependent}, the model's inputs are given by $E_{in}{=}|E_{cls}, E_v, E_a, E_t| \in \mathbb{R}^{(L+1) \times d_{g}}$, where $L{=}L_{vg} + L_{ag} + L_{tg}$ and $E_{cls}$ is a learnable CLS token.
    \item[] For \textbf{query-agnostic}, the model's inputs are given by $E_{in}{=}|E_{cls}, E_v, E_a| \in \mathbb{R}^{(L+1) \times d_{g}}$, where $L{=}L_{vg} + L_{ag} $ and $E_{cls}$ is a learnable CLS token.
\end{itemize}


A modality-specific positional embedding is also added to each input tensor; specifically, we adopt a sinusoidal positional embedding~\cite{NIPS2017_attent} for the video modality, while we find that learnable positional embeddings~\cite{li2021learnable} work best with the audio and text modalities.\\

\noindent\textbf{Architecture.}
We adopt a transformer encoder~\cite{moment-detr,Kamath2021MDETRM} which is a powerful yet flexible architecture for processing sequential data. The input $E_{in}$ is fed to a stack of $L_{t}$ transformer encoder layers. Each transformer encoder layer is identical in design to~\cite{NIPS2017_attent,Yao2021EfficientDI,moment-detr,Kamath2021MDETRM}, using a multi-head self-attention layer and a feed-forward network (FFN). 
Finally, the first element of the encoder output $(E_{out})$, corresponding to the CLS token position, is fed to an MLP for predicting the plausibility score for the given window.\\  

\noindent\textbf{Loss function and supervision definition.}
We optimize a binary cross entropy (BCE) loss to train $\mathcal{G}$. 
The query-agnostic models are trained on a dataset $\mathcal{D}_{agnostic}$ that contains only a set of windows and a binary label that indicates, for each window, whether or not it contains at least one moment. The query-dependent variant is trained on a dataset $\mathcal{D}_{dependent}$ that contains window and query pairs making up the set of describable (positives) and negative windows.

\section{Experiments}
\label{sec:experiments}
\noindent\textbf{Metrics.} The video grounding metric of choice is Recall@$K$ for IoU=$\theta$ (R@$K$-IoU=$\theta$). This widely adopted metric measures the quality of the predictions' ranking and temporal overlapping (IoU) with the annotations. In this work we evaluate our models for $K{\in}\{1, 5, 10, 50, 100\}$ and $\theta{\in}\{0.1, 0.3, 0.5\}$.
Additionally, we introduce Mean Recall@$K$ (mR@$K$), where we average the R@$K$-IoU=$\theta$ performance across all IoUs thresholds. This metric allows for easier comparison and more compact tables. We also include Mean Recall$_{all}$ or average Recall as: mR$_{all} =\frac{1}{|K|}\sum$ mR@$K$, $\forall K$.\\

\noindent\textbf{Datasets.}
MAD~\cite{Soldan_2022_CVPR} is a large-scale dataset for the video-language grounding task that comprises more than $384$K natural queries temporally grounded in $650$ full-length movies for a total of over $1.2$K hours of video. MAD introduces a new long-form video grounding setup that brings new challenges to the task at hand while allowing for unprecedented bias-free performance investigation. In our experimental section, unless otherwise specified, we report performance on the official test set. 
\noindent Ego4D~\cite{ego4d} is a comprehensive benchmark consisting of egocentric videos from 931 camera wearers worldwide in different scenarios. The subtask we focus on is the Natural Language Query (NLQ), which uses 13 question types to locate different types of information and retrieve relevant moments from the episodic memory of camera wearers. Videos vary in length from 3.5 to 20 minutes.\\

\noindent\textbf{Baselines.}
For MAD dataset, we select three video grounding methods for our experimental setup, namely:
VLG-Net~\cite{soldan2021vlg}, zero-shot CLIP~\cite{Soldan_2022_CVPR}, and Moment-DETR~\cite{moment-detr}. The first two are both proposal-based methods, a trained method and a zero-shot one, while the third is a proposal-free approach. On the other hand, for Ego4D, we use Moment-DETR~\cite{moment-detr} and VSL-Net~\cite{vslnet}, a multimodal span-based framework based on context-query attention. We will show that all approaches benefit from the score refinements provided by our Guidance Model. 

\subsection{Implementation details}\label{sub:imple}
\noindent\textbf{Feature Extraction.}
The visual and text embeddings are extracted following the CLIP-based methodology presented in~\cite{Soldan_2022_CVPR}. Specifically, visual features are extracted at $5$ FPS and with embedding dimensionality $D_v{=}512$. On the other hand, text features are comprised of $N$ tokens, with embedding dimensionality $D_t{=}512$. Finally, audio embeddings are computed using OpenL3~\cite{8682475}, an audio embedding model inspired by L3-Net~\cite{8237335}. Specifically, we use the OpenL3 checkpoint that was pre-trained on videos containing environmental audiovisual data, we use a spectrogram time-frequency representation with $128$ bands, and we set the audio embedding dimensionality $D_a$ to $512$. Furthermore, we extract the audio embeddings using a stride size equal to $0.2$ seconds, \textit{i.e.}, using an extraction frame rate of $5$ Hz, which matches the frame rate of the visual features. On the other hand, in the Ego4D experiment we follow~\cite{egovlp} without using audio streams, as not all videos contain audio throughout their entire duration.\\

\begin{table*}[!t]
\centering
\setlength{\tabcolsep}{2.5pt}
\renewcommand{\arraystretch}{1} 
\resizebox{\linewidth}{!}{%
\footnotesize
\begin{tabular}{@{}c@{\hspace{0.4em}} 
l 
c@{\hspace{0.2em}} 
ccccc c@{\hspace{0.2em}} 
ccccc c@{\hspace{0.2em}} 
cccc  c@{}}

\toprule
& \multirow{2}{*}{\textbf{Model}}
& \phantom{} & \multicolumn{5}{c}{\textbf{IoU=0.1}} 
& \phantom{} & \multicolumn{5}{c}{\textbf{IoU=0.3}}
& \phantom{} & \multicolumn{5}{c}{\textbf{IoU=0.5}} \\

\cmidrule{4-8} \cmidrule{10-14} \cmidrule{16-20} 
&
&& \textbf{R@1} & \textbf{R@5} & \textbf{R@10} & \textbf{R@50} & \textbf{R@100}
&& \textbf{R@1} & \textbf{R@5} & \textbf{R@10} & \textbf{R@50} & \textbf{R@100}
&& \textbf{R@1} & \textbf{R@5} & \textbf{R@10} & \textbf{R@50} & \textbf{R@100} \\
\midrule
          
&Zero-shot CLIP~\cite{Soldan_2022_CVPR}  
    && $6.57$ & $15.05$ & $20.26$ & $37.92$ & $47.73$ 
    && $3.13$ & $9.85$  & $14.13$ & $28.71$ & $36.98$ 
    && $1.39$ & $5.44$  &  $8.38$ & $18.80$ & $24.99$  \\
                                  
&VLG-Net~\cite{soldan2021vlg}     
    && $3.50$ & $11.74$ & $18.32$ & $38.41$ & $49.65$ 
    && $2.63$ & $9.49$  & $15.20$ & $33.68$ & $43.95$ 
    && $1.61$ & $6.23$  & $10.18$ & $25.33$ & $34.18$  \\
                                  
&Moment-DETR~\cite{moment-detr}   
    && $0.31$ & $1.52$ & $2.79$ & $11.08$ & $19.65$ 
    && $0.24$ & $1.14$ & $2.06$ & $7.97$  & $14.29$ 
    && $0.16$ & $0.68$ & $1.20$ & $4.71$  & $8.46$     \\

\cmidrule{1-20}
                                  
&$\dagger$Zero-shot CLIP~\cite{Soldan_2022_CVPR} && 
    $\mathbf{9.30}$ & $\mathbf{18.96}$ & $24.30$ & $39.79$ & $47.35$ 
    && $\mathbf{4.65}$ & $13.06$ & $17.73$ & $32.23$ & $39.58$ 
    && $2.16$ & $7.40$ & $11.09$ & $23.21$ & $29.68$   \\

&$\dagger$VLG-Net~\cite{soldan2021vlg}    
    && $5.60$ & $16.07$ & $23.64$ & $45.35$ & $55.59$ 
    && $4.28$ & $\mathbf{13.14}$ & $\mathbf{19.86}$ & $39.77$ & $49.38$ 
    && $\mathbf{2.48}$ & $\mathbf{8.78}$ & $\mathbf{13.72}$ & $\mathbf{30.22}$ & $\mathbf{39.12}$  \\

&$\dagger$Moment-DETR~\cite{moment-detr} 
    && $5.07$ & $16.30$ & $\mathbf{24.79}$ & $\mathbf{50.06}$ & $\mathbf{61.79}$ 
    && $3.82$ & $12.60$ & $19.43$ & $\mathbf{40.52}$ & $\mathbf{50.35}$ 
    && $2.39$ & $7.90$ & $12.06$ & $24.87$ & $30.81$  \\

\bottomrule
\end{tabular}%
}
\caption{\small\label{tab:mad_grounding}{\bf Benchmarking of grounding methods on the MAD dataset.}  
In the initial rows (row 1, 2, and 3), we present the performance results of the three baselines (Zero-shot CLIP, VLG-Net, and Moment-DETR) on the test split. The subsequent rows (row 4, 5, and 6) demonstrate the performance enhancement achieved by the grounding models aided by the proposed guidance model (indicated by the  ($\dagger$) symbol), utilizing the query-dependent setup with audiovisual features. 
}
\end{table*}

\begin{table}[t]
\centering
\setlength{\tabcolsep}{2.5pt}
\renewcommand{\arraystretch}{1} 
\resizebox{\columnwidth}{!}{%
\footnotesize
\begin{tabular}{@{}lllllllll@{}}
\toprule
\multicolumn{3}{l}{\textbf{Method}} & \textbf{mR@1} & \textbf{mR@5} & \textbf{mR@10} & \textbf{mR@50} & \textbf{mR@100} & \textbf{mR$_{all}$} \\ \midrule
\multicolumn{3}{l}{Moment-DETR~\cite{moment-detr}} & $6.72$ & $19.68$ & $23.85$ & $24.67$ & $-$ & $18.73$\\
\multicolumn{3}{l}{VSL-Net~\cite{vslnet,egovlp}} & $11.38$ & $19.64$ & $24.27$ & $36.09$ & $42.12$ & $26.70$\\

\cmidrule{1-9} 
\multicolumn{3}{l}{$\dagger$Moment-DETR~\cite{moment-detr}} & $7.28$ & $22.14$ & $24.75$ & $26.05$ & $-$ & $20.05$\\
\multicolumn{3}{l}{$\dagger$VSL-Net~\cite{vslnet,egovlp}} & $\mathbf{11.90}$ & $\mathbf{24.06}$ & $\mathbf{33.31}$ & $\mathbf{54.59}$ & $\mathbf{54.90}$ & $\mathbf{31.22}$ \\
\bottomrule
\end{tabular}
}
\caption{\small\label{tab:egod4}{\bf Benchmarking on the Ego4D dataset.} 
The initial rows (row 1 and 2) display the performance results of Moment-DETR~\cite{moment-detr} and VSL-Net~\cite{vslnet} baselines, employing Egocentric VLP features~\cite{egovlp} on the Ego4D validation set. The subsequent rows (row 3 and 4) demonstrate the performance enhancement brought about by our Guidance Model, indicated by the symbol $\dagger$. For a complete table of results, kindly consult the supplementary material.
}
\end{table}

\noindent\textbf{Guidance Model.} We train our Guidance Model using three modalities: \textbf{(i)} visual, \textbf{(ii)} audio, and \textbf{(iii)} text for \textit{query-dependent} setup. And two modalities: \textbf{(i)} visual and \textbf{(ii)} audio for \textit{query-agnostic} setup. Our transformer encoder comprises $6$ layers $(L_t)$ with a hidden size of $256$ $(d_g)$. The Guidance Model is trained using a sliding window approach where the window size $(L_{vg})$ is equal to $64$ frames (unless otherwise specified), spanning $12.8$ seconds for the MAD dataset and $34.13$ seconds for the Ego4D dataset. The model is trained for $100$ epochs using the AdamW~\cite{adamw} optimizer with learning rate $10^{-4}$, weight decay $10^{-4}$, and batch size $512$. \\

\noindent\textbf{Grounding Models.} We instantiate Moment-DETR~\cite{moment-detr} using a hidden dimension size of $256$, $2$ encoder layers, $2$ decoder layers, window length $(L_v)$ of $128$ (equivalent to $25.6$ seconds at $5$ FPS in MAD), and $10$ moment queries; we optimize the model with AdamW~\cite{adamw}, setting the learning rate to $10^{-4}$ and the batch size to $256$. VLG-Net~\cite{soldan2021vlg} is trained following the implementation details presented in~\cite{Soldan_2022_CVPR}, while CLIP is evaluated in the zero-shot setup that was also proposed in~\cite{Soldan_2022_CVPR}. For VSL-Net~\cite{vslnet}, we follow the approach presented in the Ego4D~\cite{ego4d} repository with an egocentric video-language pretraining~\cite{egovlp}. At inference time, we discard highly redundant proposals via non-maximum suppression (NMS) with a threshold of $0.3$ for MAD and $0.5$ for Ego4D. All experiments are conducted on a Linux workstation with a single NVIDIA 32GB V$100$ GPU and an Intel Xeon CPU with $64$ cores.

\subsection{Results}
Table~\ref{tab:mad_grounding} presents the performance of the baselines (rows 1-3) and the corresponding improvement achieved by the grounding models with the guidance model (rows 4-6). The employed guidance model utilizes query-dependent setup and audiovisual features. Further investigation into various model design choices can be found in Section~\ref{sec:ablation}.


\noindent As anticipated, proposal-based methods (rows 1-2) have consistently higher performance with respect to the proposal-free Moment-DETR method (row 3) for all metrics. Moreover, our guidance method is able to boost the performance of all baselines (rows 4-6). In particular, the most significant boost is obtained by Moment-DETR, for which the improvement ranges between $3{-}16\times$, while for VLG-Net and zero-shot CLIP baselines, the improvement ranges between $1{-}3\times$. Remarkably, the Guidance Model can boost R@$10$-IoU=$0.1$ from $2.79\%$ to $24.79\%$ for Moment-DETR, allowing us to achieve state-of-the-art performance for the MAD dataset. 
These results indicate that our approach bridges the gap between proposal-free and proposal-based methods, allowing the former to tackle the long-form video grounding task.

We evaluated the performance of the guidance model on the Ego4D~\cite{ego4d} dataset using a query-dependent setup. Since Ego4D lacks audio in a substantial portion of its videos for the NLQ task, we opted for a fair comparison using only visual and text features. Table~\ref{tab:egod4} displays the results achieved, employing VSL-Net and Moment-DETR as the baselines. The results demonstrate that the Guidance Model had a positive impact, leading to a $4.52\%$ improvement in the mR$_{all}$ metric for VSL-Net and a $1.32\%$ improvement for Moment-DETR in the same metric.



\noindent \textbf{Takeaway.} Our Guidance Model is general and can be combined with all grounding methods to bridge their performance from the short-form to the long-form setup. We refer the reader to the supplementary material for additional evaluations of the baselines on the short-form video grounding setup.


\subsection{Ablation Study}\label{sec:ablation}
In order to determine the optimal configuration of the Guidance Model, we conducted ablation studies focusing on three critical factors: \textbf{(i)} the selection of modalities (i.e., visual and/or audio), \textbf{(ii)} the performance of query-agnostic versus query-dependent guidance, and \textbf{(iii)} determining the optimal window size using the MAD dataset. Moreover, we performed an extensive investigation on actionless moments, which provides a significant differentiation of our implementation from the temporal proposals method. Moreover, we investigated the influence of the audio streams in video grounding models. Lastly, we conclude with a qualitative analysis of our proposed pipeline. Full detailed results in supplementary material. \\



\noindent\textbf{Guidance through multimodality fusion.}
In this section, we investigate the contribution of having multiple modalities for the Guidance Model. In Table~\ref{tab:multi-mod} we report the baselines' performance as-is (rows 1,5,9), as well as their performances when combined with our Guidance Model. The Guidance Model in each row was trained with different input modalities for the MAD dataset: (i) audio and text, (ii) video and text, (iii) audio, video, and text. We report Mean Recall@$K$ (mR@$K$) for compactness. 

\begin{table}[!t]
\centering
\setlength{\tabcolsep}{2.5pt}
\renewcommand{\arraystretch}{1} 
\resizebox{\columnwidth}{!}{%
\footnotesize
\centering
\begin{tabular}{lccccccc}

\toprule
\multirow{2}{*}{\textbf{Model}} & 
\multicolumn{2}{c}{\textbf{Modalities}}  & 
\multirow{2}{*}{\textbf{mR@1}}  &
\multirow{2}{*}{\textbf{mR@5}}  &
\multirow{2}{*}{\textbf{mR@10}} &
\multirow{2}{*}{\textbf{mR@50}} &
\multirow{2}{*}{\textbf{mR@100}}  \\ 

& \textbf{Audio} & \textbf{Visual} \\ 
\hline

\multirow{4}{1cm}{Zero-shot CLIP}  
& \xmark & \xmark & $3.7$ & $10.0$ & $13.9$ & $27.3$ & $35.0$ \\
& \cmark & \xmark & $4.0$ & $11.1$  & $15.6$ & $28.1$ & $35.3$ \\ 
& \xmark & \cmark & $4.8$ & $12.0$ & $16.1$ & $29.2$ & $36.0$ \\ 
& \cmark & \cmark & $\mathbf{5.2}$ & $\mathbf{12.6}$ & $\mathbf{16.7}$ & $\mathbf{29.7}$  & $\mathbf{36.4}$\\ 
\cmidrule{1-8}

\multirow{4}{1cm}{VLG-Net}
& \xmark & \xmark & $2.5$ & $8.6$ & $13.4$ & $31.0$ & $40.8$\\ 
& \cmark & \xmark & $2.7$ & $9.2$ & $14.1$ & $32.1$ & $41.7$\\ 
& \xmark & \cmark & $3.5$ & $11.1$ & $16.7$ & $35.0$ & $44.1$\\ 
& \cmark & \cmark & $\mathbf{3.9}$ & $\mathbf{12.1}$ & $\mathbf{17.8}$ & $\mathbf{36.0}$  & $\mathbf{45.2}$\\ 
\cmidrule{1-8}

\multirow{4}{1cm}{Moment-DETR}
& \xmark & \xmark & $0.2$ & $1.0$ & $1.8$  & $7.5$  &  $13.1$\\ 
& \cmark & \xmark & $1.0$ & $4.2$ & $7.1$  & $20.7$ &  $29.5$\\ 
& \xmark & \cmark & $3.1$ & $10.7$ & $16.2$ & $34.1$ &  $42.9$\\ 
& \cmark & \cmark & $\mathbf{3.6}$ & $\mathbf{11.5}$ & $\mathbf{17.2}$ & $\mathbf{35.2}$ & $\mathbf{43.8}$\\ 
\bottomrule
\end{tabular}%
}
\caption{\small\label{tab:multi-mod}{\bf Modality comparison for the Query Dependent Guidance Model.} The table reports the boost in performance achieved by the chosen baselines (rows 1,5,9) when combined with a Guidance Model having access to different input modalities. Mean Recall (mR@K) metric is computed over the validation set. }
\end{table}

For each baseline, the best performance can be achieved when all modalities are used for the Guidance Model, yielding an improvement between $1{-}18\times$ across all metrics. 
Using audio alone without visual input leads to modest performance improvement for Zero-shot CLIP, VLG-Net, and Moment-DETR. Additionally, visual clues can also boost all baselines, with Moment-DETR benefiting the most. Nonetheless, when combining all modalities, we can achieve the best overall performance. 

Notice how the use of the Guidance Model is able to boost the performance of Moment-DETR, getting it to achieve a performance close to the proposal-based method even though the baseline performance is particularly poor. This finding allows us to conclude that strong short-video grounding methods can be bridged to long-form grounding through a two-stage approach that leverages a Guidance Model to reduce the search space for the temporal alignment of video and text. \\

\noindent\textbf{Describable windows.}
\label{ablation:describabe}
The first ablation suggests that using both audio and video is a powerful representation. However, the Guidance Model used so far is query-dependent and leverages textual queries for identifying irrelevant windows. Our method can be very efficient if we can identify windows that are non-describable regardless of the input query so that the Guidance Model can process the video/audio streams only once. 
The question we want to answer in the following ablation is \textit{``Can we devise an efficient first stage to identify non-describable windows and remove them from the search space of grounding methods?''} 
To investigate this research question, we devise a query-agnostic Guidance Model that does not process any textual query in input. In Table~\ref{tab:query-de}, we report the comparison between the query-agnostic and query-dependent Guidance Model, which uses both audio and visual cues as inputs. For compactness, we report the Mean Recall$_{all}$ (mR$_{all}$).
\noindent Using audio-visual cues alone leads to improvements for Zero-shot CLIP, VLG-Net, and Moment-DETR. Conversely, when our Guidance Model takes the text query as input, it can better discriminate between relevant and irrelevant moments leading to consistent improvements across all baselines. We conclude that the query-dependent setup offers superior performance, at the cost of being more computationally
expensive as the number of queries grows. On the other hand, the query-agnostic setup is computationally efficient as the video only has to be processed once by the Guidance Model, making it suitable for real-time or low-resource scenarios.\\

\begin{table}[!t]
\centering
\setlength{\tabcolsep}{2.5pt}
\renewcommand{\arraystretch}{1} 
\resizebox{.9\columnwidth}{!}{%
\footnotesize
\begin{tabular}{lccccc}
\toprule
\multirow{2}{*}{\textbf{Model}}      & 
\multirow{2}{*}{\textbf{Baseline}}   & 
\multicolumn{2}{c}{\textbf{Query}}   & 
\multicolumn{2}{c}{\textbf{Query}}   \\

&& \multicolumn{2}{c}{\textbf{Agnostic}} & \multicolumn{2}{c}{\textbf{Dependent}} \\ 
\hline
Zero-shot CLIP~\cite{Soldan_2022_CVPR}     & $18.0$ & $18.5$ & ${\scriptscriptstyle(+0.5)}$ & $\mathbf{20.1}$ & ${\scriptscriptstyle(+2.1)}$  \\ 
VLG-Net~\cite{soldan2021vlg}            & $19.3$ & $20.7$ & ${\scriptscriptstyle(+1.4)}$ & $\mathbf{23.0}$ & ${\scriptscriptstyle(+3.7)}$  \\ 
Moment-DETR~\cite{moment-detr}      &  $4.7$ &  $8.3$ & ${\scriptscriptstyle(+3.6)}$ & $\mathbf{22.3}$ & ${\scriptscriptstyle(+17.6)}$ \\ 
\bottomrule
\end{tabular}%
}
\caption{\small\label{tab:query-de}{\bf Describable windows.} We report Mean Recall$_{all}$ for the baselines performances (second column), query agnostic guidance filter combined with baselines (third column), and query dependent guidance filter combined with baselines (fourth column). Performance is measured on the MAD validation set. }
\end{table}

\begin{figure}[t]
    \centering
    \includegraphics[width=\linewidth]{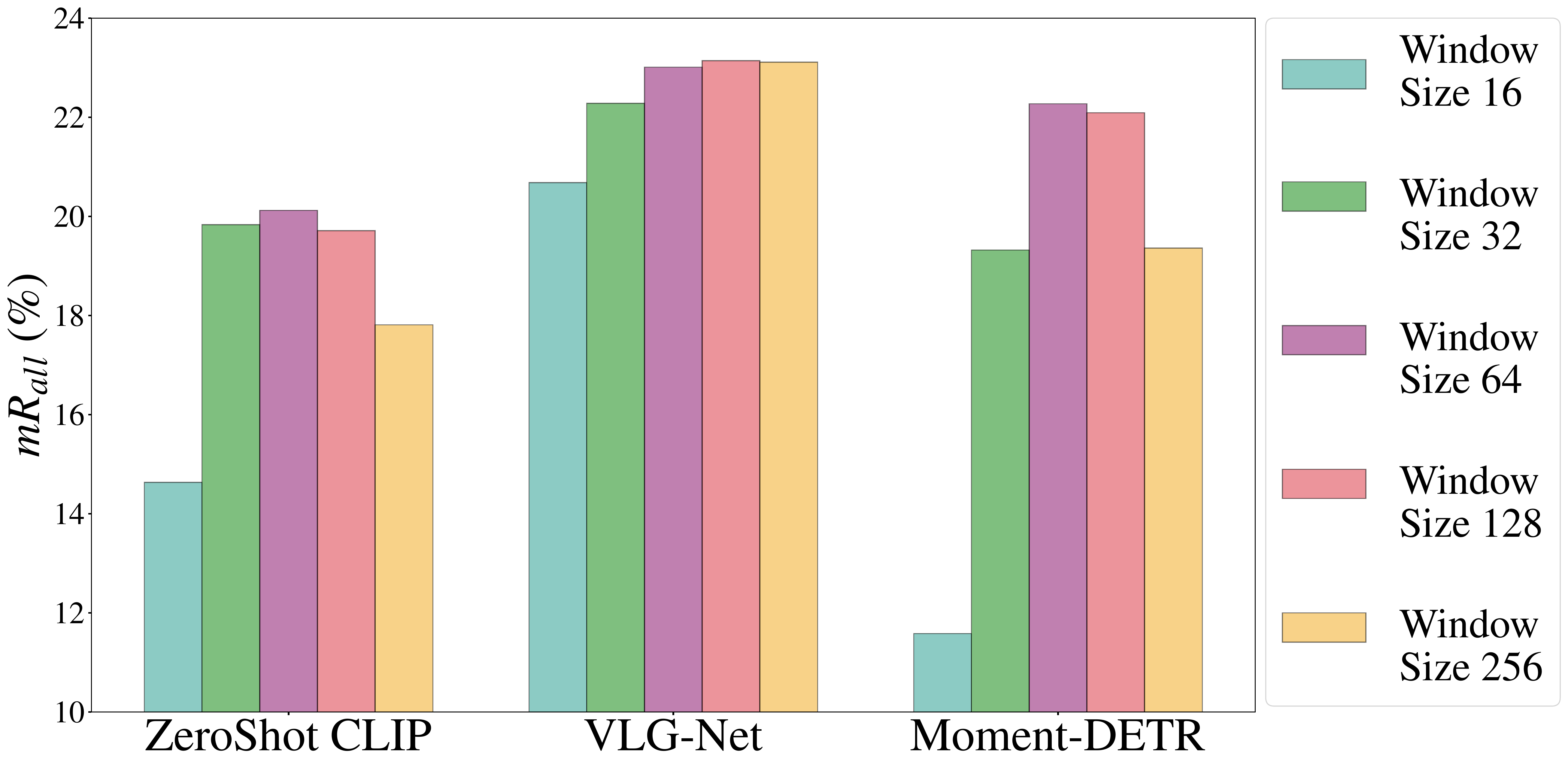}
    \small\caption{\small\textbf{Temporal field of view.} 
    We report mR$_{all}$ for the combination of three baseline models (Zero-shot CLIP, VLG-Net, and Moment-DETR) with different \textit{Guidance Models} trained with different window sizes. Results are computed on the validation split. 
    }
    \label{fig:model-filters}
    \vspace{-.4cm}
\end{figure}

\noindent\textbf{Temporal field of view.}
The next question we want to answer is: \textit{``What is the optimal window size for the Guidance Model?''}. 
In Figure~\ref{fig:model-filters}, we report the performance trend for all baselines when combined with a query-dependent audio-visual Guidance Model, where we vary the window size (temporal field of view) the Guidance Model can reason about. The trade-off in this experiment is between the fine-grained guidance versus the temporal context available to the guidance model. We test window sizes ranging from $16$ to $256$ time steps in the MAD dataset, with each time step accounting for $0.2$s when the video FPS is set to $5$. This yields models with a temporal field of view ranging from $3.2$ to $51.2$ seconds. For this experiment, we report the mR$_{all}$ metric. 

Figure \ref{fig:model-filters} showcases that, although short windows provide a more fine-grained guidance, they do not yield the best performance. For all methods investigated, the performance consistently increases from windows size $16$ to $64$.
After such a window size, performance starts to decrease for the Zero-Shot CLIP and Moment-DETR, while little change is observed for VLG-Net. Following these findings, we fix the window size for the guidance model to be $64$ in all other experiments. 
 Hence, we can conclude that to improve the performance of the video grounding task, it is necessary to have sufficient context information so as to delimit all possible predictions correctly. \\ 

\begin{table}[t]

\centering
\setlength{\tabcolsep}{2.5pt}
\renewcommand{\arraystretch}{1} 
\resizebox{.70\columnwidth}{!}{%
\footnotesize
\begin{tabular}{@{}lccccc@{}}
\toprule
\multicolumn{3}{l}{\textbf{Method}} & \textbf{mR@1} & \textbf{mR@5} & \textbf{mR@10} \\ 

\midrule

\multicolumn{3}{l}{Zero-shot CLIP~\cite{Soldan_2022_CVPR}} 
& $3.09$ & $8.10$ & $11.30$ \\

\multicolumn{3}{l}{VLG-Net~\cite{soldan2021vlg}} 
& $1.53$ & $5.77$ & $9.33$ \\

\multicolumn{3}{l}{Moment-DETR~\cite{moment-detr}  } 
& $0.15$ & $0.65$ & $0.91$ \\

\cmidrule{1-6}

\multicolumn{3}{l}{$\dagger$Zero-shot CLIP} 
& $4.31$ & $10.38$ & $14.01$ \\

\multicolumn{3}{l}{$\dagger$VLG-Net} 
& $2.32$ & $8.07$ & $13.13$ \\

\multicolumn{3}{l}{$\dagger$ Moment-DETR} 
& $2.86$ & $8.44$ & $13.51$ \\ 

\bottomrule
\end{tabular}
}
\caption{\small\label{tab:non-action}{\bf Describable windows beyond actions.} We compare the performance of our baseline models with the Guidance Model ($\dagger$) and without guidance on actionless queries only. For this purpose, we extract actionless queries from MAD~\cite{Soldan_2022_CVPR} test set and evaluate using mR@$K$ for $K{\in}\{1, 5, 10\}$. The results showcase that  our guidance method is not solely learning action-based concepts.  }
\vspace{-.3cm}
\end{table}

\noindent\textbf{Describable windows beyond actions.}
Our motivation is to follow a data-driven approach to learn about what segment of the videos are worth describing. Looking at the data in MAD~\cite{Soldan_2022_CVPR} reveals that a considerable portion of the queries (10\% and 18\%, respectively) lack verbs and describe only the environment and its attributes (adjectives, nouns). Examples of such queries include \textit{``Night time at SOMEONE's building"} and \textit{``Later, on a circular staircase"}. We recognize that interpreting this type of query is a challenging task, and it sets our approach apart from other computer vision methods, such as temporal proposals~\cite{SST,SSN2017ICCV,AAAIYangWW0XYH22,HAN2022217,qing2021temporal} because these moments do not contain any action element. Therefore, to validate the above hypothesis, we report in Table~\ref{tab:non-action} the boost in performance that our Guidance Model brings for actionless queries. The metrics show an improvement when our method is applied to actionless queries, providing evidence for the effectiveness of our approach in capturing descriptions of actionless moments.\\


\noindent\textbf{Enhancing Video Grounding by Using Audio.} By incorporating audio features in Moment-DETR~\cite{moment-detr} using the MAD Dataset~\cite{Soldan_2022_CVPR}, we achieve a substantial performance boost. As shown in Table~\ref{tab:ground_audio}, the mR${all}$ (mean Recall over all classes) improves from $5.08\%$ to $7.70\%$. Furthermore, when combined with the Guidance Model, the mR${all}$ further increases from $24.19\%$ to $28.30\%$, demonstrating a consistent impact across setups. These results show a remarkable performance improvement when audio is integrated into the video grounding model. 

We emphasize that our main focus is not on adding audio to the video grounding model but rather on the design and experimentation of the Guidance model. For a fair comparison, models in Tables~\ref{tab:mad_grounding} and~\ref{tab:egod4} solely use visual and language features, excluding audio. However, we believe these results show encouraging results for the future inclusion of audio features in mainstream grounding models. \\

\begin{table}[t]
\vspace{.1cm}
\centering
\setlength{\tabcolsep}{1.5pt}
\renewcommand{\arraystretch}{1} 
\resizebox{1\columnwidth}{!}{%
\footnotesize
\begin{tabular}{@{}cccccccc@{}}
\toprule
\textbf{Method}  & \textbf{ Audio } & \textbf{m@R1} & \textbf{m@R5}  & \textbf{m@R10} & \textbf{m@R50}  & \textbf{m@R100}  & \textbf{mR$_{all}$}   \\ \midrule
Moment-DETR~\cite{moment-detr}  & \xmark  & $0.24$ & $1.11$ & $2.02$  & $7.92$  & $14.13$ & $5.08$  \\
Moment-DETR~\cite{moment-detr}  & \cmark  & $0.70$ & $2.15$  & $5.11$  & $11.23$ & $19.08$ & $7.70$  \\ \midrule
$\dagger$Moment-DETR~\cite{moment-detr}  & \xmark  & $3.76$ & $12.27$ & $18.78$ & $38.48$ & $47.65$ & $24.19$ \\
$\dagger$Moment-DETR~\cite{moment-detr}  & \cmark & $\mathbf{5.18}$ & $\mathbf{15.70}$ & $\mathbf{22.04}$ & $\mathbf{44.26}$ & $\mathbf{54.32}$ & $\mathbf{28.30}$ \\ \bottomrule
\end{tabular}
}
\caption{\small{\label{tab:ground_audio}{\bf Including audio in Grounding Model.} 
The results demonstrate a performance boost when audio is integrated into the video grounding model, thereby opening new possibilities for future research to explore this modality further. We enhance reader understanding by emphasizing methods that include the Guidance Model with ($\dagger$) in their name.}}
\end{table}


\begin{figure*}[ht]
    \centering
    \includegraphics[width=0.88\textwidth]{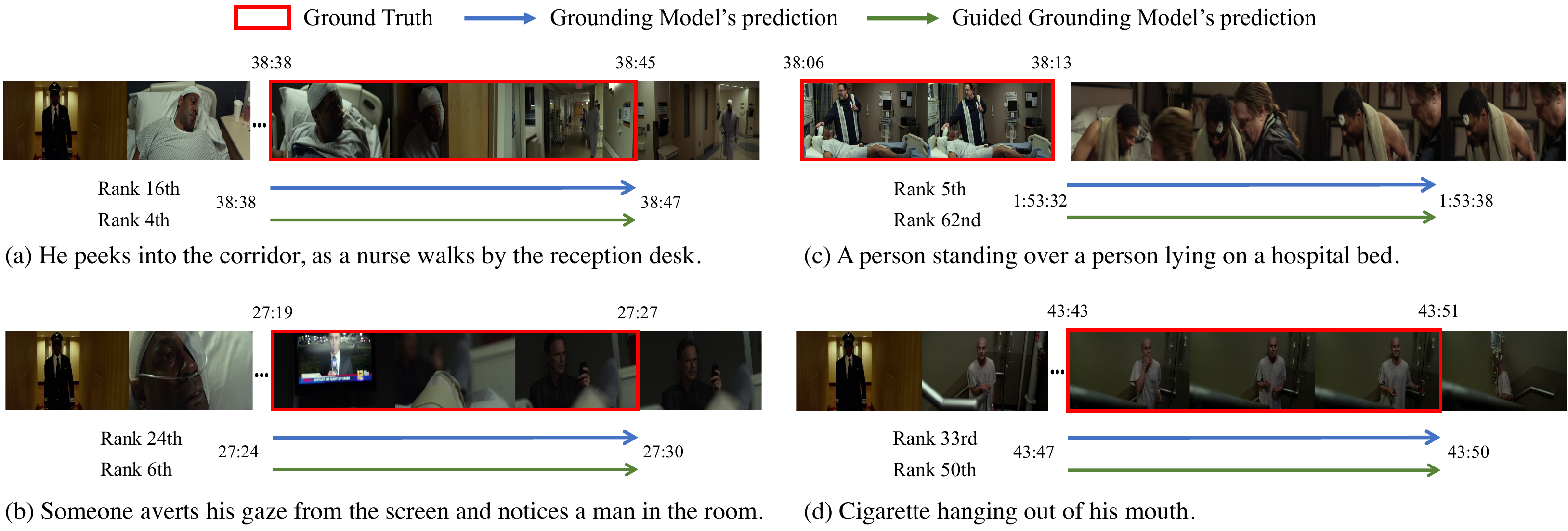}
    \caption{\small\textbf{Qualitative Results.} We compare ground truth annotation (red box) and the predicted temporal endpoints (arrows) with their respective rankings. We highlight in blue the prediction from the baseline VLG-Net~\cite{soldan2021vlg} and in green the prediction of the baseline when combined with our guidance model. 
    Notably, examples (a-c) depict a positive impact of the guidance model over the baseline one by either improving the ranking of a prediction or worsening it. Example (d) instead depicts a failure case for our pipeline.
    } 
    \label{fig:visualization}
    \vspace{-0.35cm}
\end{figure*}


\noindent\textbf{Qualitative Results.}
Figure~\ref{fig:visualization} presents three qualitative grounding results from the MAD~\cite{Soldan_2022_CVPR} test set. The figure showcases how our Guidance Model can improve the ranking of moments predicted by a grounding model (\textit{i.e.}, VLG-Net~\cite{soldan2021vlg}).
We report three positive cases (a-c) and a failure case (d).
In Figure \ref{fig:visualization}a, the Guidance Model is able to improve the ranking of the baseline model's prediction by $12$ positions (from rank $16$ to rank $4$). Given the tight tIoU between the predicted moment and the ground truth, the Guidance Model is able to positively boost the performance. An even larger improvement is showcased in \ref{fig:visualization}b, where the Guidance Model pushes the baseline prediction from rank $24$ to $6$.
Figure~\ref{fig:visualization}c depicts a case when our Guidance Model helps to filter out a non-describable window by sensibly reducing its ranking from $5$ to $62$. 
Here, the moment predicted by the baseline model (blue) is ranked high, however, the predicted moment possesses the characteristics of a non-describable window. Therefore, our Guidance Model penalizes such a moment by reducing its ranking, making this false positive less relevant. Combined with our experimental section, the visualization provides a clearer understanding of the function of our guidance filter. 
Although the Guidance Model has an overall beneficial impact over all baselines, it can also impair performance in some cases. Figure \ref{fig:visualization}d is such an example, where our Guidance Model worsens the ranking of a positive prediction; the reader should notice how this example depicts a static scene containing a moment relevant to the audience, with the Guidance Model penalizing it in a wrong way.

\section{Limitations}\label{sec:limitations}

Our method presents a notable improvement in video grounding performance. Nevertheless, it also presents a significant limitation regarding to the substantial inference time needed to match each query with every segment in a provided video. In our pursuit of finding a middle ground between computational efficiency and performance, we offer a more cost-effective option by introducing a query-agnostic framework. It is important to recognize that opting for this alternative comes with a drawback: a decrease in performance across various metrics. 
This trade-off arises because the approach shift from query-dependent framework to detecting potentially interesting moments without considering a natural language query. Nonetheless, we believe this work will inspire the community in pursuing further effective and efficient video grounding on long-form video datasets. 


\section{Social impact}\label{sec:social}
Our models development heavily leveraged the MAD dataset~\cite{Soldan_2022_CVPR}. This dataset, primarily drawn from cinematic sources, underpins our training efforts but also introduces the biases inherent in movie portrayals. While providing benefits for training, this method also brings in biases come from historical stories, cultural portrayals, and societal norms seen in movies. These biases could affect the model's comprehension of real-world concepts. To address this, we conducted experiments using the Ego4D dataset~\cite{ego4d}. This dataset takes a different approach by collecting videos worldwide in partnership with local organizations. Its goal is to capture a variety of everyday experiences across different cultures and areas, in order to offset any biases that may arise from how these experiences are depicted in movies. Impressively, our models demonstrated robust performance across both MAD and Ego4D datasets, showcasing their versatility and suggesting their potential for knowledge transfer. Moreover, the incorporation of Ego4D experiments contribute to the broader discourse on fair AI by promoting testing video-based approaches on such diverse datasets.

\section{Conclusion}\label{sec: conclusions}
We presented a novel approach for language grounding in long videos. Our guidance framework is flexible, and our experimental section provides ample demonstration that it can boost performance for a variety of baselines, including proposal-based models like VLG-Net and zero-shot CLIP, as well as  VSL-Net, and the proposal-free Moment-DETR.
Our proposed Guidance Model can operate over different modalities and is very efficient in its query-agnostic version. However, we found that using the queries as part of the guidance mechanism allows for better performance. Future work includes exploring the use of smaller, more specialized Guidance Models to optimize inference time by prioritizing recall or precision based on application needs, effectively reducing inference time while maintaining desired performance levels. This could lead to a customized processing pipeline for various applications, achieving a balance between computational efficiency and accuracy. We also believe the community will further pursue two-stage approaches for the task at hand by designing more sophisticated and powerful guidance models that can provide even better assistance to paired grounding methods.\\

\noindent\textbf{Acknowledgments.} \textit{W. Barrios} was supported by Dartmouth Graduate Fellowship through Guarini Office and Computer Science Department at Dartmouth College. 

\textit{M. Soldan} was supported by the King Abdullah University of Science and Technology (KAUST) Office of Sponsored Research through the Visual Computing Center (VCC) funding, as well as, the SDAIA-KAUST Center of Excellence in Data Science and Artificial Intelligence (SDAIA-KAUST AI). 

\textit{ A. Ceballos Arroyo} acknowledges financial support for this work by the Fulbright U.S. Student Program, which is sponsored by the U.S. Department of State and Fulbright Colombia. In addition, he was partially supported by Northeastern University through a Graduate Fellowship.

{\small
\bibliographystyle{ieee_fullname}
\bibliography{egbib}

\begin{thebibliography}{10}\itemsep=-1pt

\bibitem{Hendricks_2017_ICCV}
Lisa Anne~Hendricks, Oliver Wang, Eli Shechtman, Josef Sivic, Trevor Darrell,
  and Bryan Russell.
\newblock {Localizing Moments in Video With Natural Language}.
\newblock In {\em Proceedings of the IEEE International Conference on Computer
  Vision (ICCV)}, 2017.

\bibitem{8237335}
Relja Arandjelovic and Andrew Zisserman.
\newblock Look, listen and learn.
\newblock In {\em 2017 IEEE International Conference on Computer Vision
  (ICCV)}, pages 609--617, 2017.

\bibitem{brown2020language}
Tom Brown, Benjamin Mann, Nick Ryder, Melanie Subbiah, Jared~D Kaplan, Prafulla
  Dhariwal, Arvind Neelakantan, Pranav Shyam, Girish Sastry, Amanda Askell,
  et~al.
\newblock Language models are few-shot learners.
\newblock {\em Advances in neural information processing systems},
  33:1877--1901, 2020.

\bibitem{SST}
Shyamal Buch, Victor Escorcia, Chuanqi Shen, Bernard Ghanem, and Juan~Carlos
  Niebles.
\newblock Sst: Single-stream temporal action proposals.
\newblock In {\em 2017 IEEE Conference on Computer Vision and Pattern
  Recognition (CVPR)}, pages 6373--6382, 2017.

\bibitem{carion2020end}
Nicolas Carion, Francisco Massa, Gabriel Synnaeve, Nicolas Usunier, Alexander
  Kirillov, and Sergey Zagoruyko.
\newblock End-to-end object detection with transformers.
\newblock In {\em European conference on computer vision}, pages 213--229.
  Springer, 2020.

\bibitem{chenhierarchical}
{Chen Shaoxiang, Jiang Yu-Gang}.
\newblock {Hierarchical Visual-Textual Graph for Temporal Activity Localization
  via Language}.
\newblock In {\em Proceedings of the European Conference on Computer Vision
  (ECCV)}, 2020.

\bibitem{8682475}
Jason Cramer, Ho-Hsiang Wu, Justin Salamon, and Juan~Pablo Bello.
\newblock Look, listen, and learn more: Design choices for deep audio
  embeddings.
\newblock In {\em ICASSP 2019 - 2019 IEEE International Conference on
  Acoustics, Speech and Signal Processing (ICASSP)}, pages 3852--3856, 2019.

\bibitem{bert}
Jacob Devlin, Ming{-}Wei Chang, Kenton Lee, and Kristina Toutanova.
\newblock {BERT:} pre-training of deep bidirectional transformers for language
  understanding.
\newblock In Jill Burstein, Christy Doran, and Thamar Solorio, editors, {\em
  Proceedings of the 2019 Conference of the North American Chapter of the
  Association for Computational Linguistics: Human Language Technologies,
  {NAACL-HLT} 2019, Minneapolis, MN, USA, June 2-7, 2019, Volume 1 (Long and
  Short Papers)}, pages 4171--4186. Association for Computational Linguistics,
  2019.

\bibitem{dosovitskiy2020image}
Alexey Dosovitskiy, Lucas Beyer, Alexander Kolesnikov, Dirk Weissenborn,
  Xiaohua Zhai, Thomas Unterthiner, Mostafa Dehghani, Matthias Minderer, Georg
  Heigold, Sylvain Gelly, et~al.
\newblock An image is worth 16x16 words: Transformers for image recognition at
  scale.
\newblock {\em arXiv preprint arXiv:2010.11929}, 2020.

\bibitem{escorcia2019temporal}
Victor Escorcia, Mattia Soldan, Josef Sivic, Bernard Ghanem, and Bryan~C.
  Russell.
\newblock Temporal localization of moments in video collections with natural
  language.
\newblock {\em CoRR}, abs/1907.12763, 2019.

\bibitem{Gao_2017_ICCV}
{Gao Jiyang, Sun Chen, Yang Zhenheng, Nevatia, Ram}.
\newblock {TALL: Temporal Activity Localization via Language Query}.
\newblock In {\em Proceedings of the IEEE International Conference on Computer
  Vision (ICCV)}, 2017.

\bibitem{ego4d}
Kristen Grauman, Andrew Westbury, Eugene Byrne, Zachary~Q. Chavis, Antonino
  Furnari, Rohit Girdhar, Jackson Hamburger, Hao Jiang, Miao Liu, Xingyu Liu,
  Miguel Martin, Tushar Nagarajan, Ilija Radosavovic, Santhosh~K. Ramakrishnan,
  Fiona Ryan, Jayant Sharma, Michael Wray, Mengmeng Xu, Eric~Z. Xu, Chen Zhao,
  Siddhant Bansal, Dhruv Batra, Vincent Cartillier, Sean Crane, Tien Do, Morrie
  Doulaty, Akshay Erapalli, Christoph Feichtenhofer, Adriano Fragomeni, Qichen
  Fu, Christian Fuegen, Abrham Gebreselasie, Cristina Gonz{\'a}lez, James~M.
  Hillis, Xuhua Huang, Yifei Huang, Wenqi Jia, Weslie Yu~Heng Khoo, J{\'a}chym
  Kol{\'a}r, Satwik Kottur, Anurag Kumar, Federico Landini, Chao Li, Yanghao
  Li, Zhenqiang Li, Karttikeya Mangalam, Raghava Modhugu, Jonathan Munro,
  Tullie Murrell, Takumi Nishiyasu, Will Price, Paola~Ruiz Puentes, Merey
  Ramazanova, Leda Sari, Kiran~K. Somasundaram, Audrey Southerland, Yusuke
  Sugano, Ruijie Tao, Minh Vo, Yuchen Wang, Xindi Wu, Takuma Yagi, Yunyi Zhu,
  Pablo Arbel{\'a}ez, David~J. Crandall, Dima Damen, Giovanni~Maria Farinella,
  Bernard Ghanem, Vamsi~Krishna Ithapu, C.~V. Jawahar, Hanbyul Joo, Kris
  Kitani, Haizhou Li, Richard~A. Newcombe, Aude Oliva, Hyun~Soo Park, James~M.
  Rehg, Yoichi Sato, Jianbo Shi, Mike~Zheng Shou, Antonio Torralba, Lorenzo
  Torresani, Mingfei Yan, and Jitendra Malik.
\newblock Ego4d: Around the world in 3,000 hours of egocentric video.
\newblock {\em 2022 IEEE/CVF Conference on Computer Vision and Pattern
  Recognition (CVPR)}, pages 18973--18990, 2021.

\bibitem{HAN2022217}
Tingting Han, Sicheng Zhao, Xiaoshuai Sun, and Jun Yu.
\newblock Modeling long-term video semantic distribution for temporal action
  proposal generation.
\newblock {\em Neurocomputing}, 490:217--225, 2022.

\bibitem{layernor}
Geoffrey~E. Hinton, Nitish Srivastava, Alex Krizhevsky, Ilya Sutskever, and
  Ruslan Salakhutdinov.
\newblock Improving neural networks by preventing co-adaptation of feature
  detectors.
\newblock {\em ArXiv}, abs/1207.0580, 2012.

\bibitem{Kamath2021MDETRM}
Aishwarya Kamath, Mannat Singh, Yann LeCun, Ishan Misra, Gabriel Synnaeve, and
  Nicolas Carion.
\newblock Mdetr - modulated detection for end-to-end multi-modal understanding.
\newblock {\em 2021 IEEE/CVF International Conference on Computer Vision
  (ICCV)}, pages 1760--1770, 2021.

\bibitem{Krishna_2017_ICCV}
Ranjay Krishna, Kenji Hata, Frederic Ren, Li Fei-Fei, and Juan Carlos~Niebles.
\newblock {Dense-Captioning Events in Videos}.
\newblock In {\em Proceedings of the IEEE International Conference on Computer
  Vision (ICCV)}, 2017.

\bibitem{moment-detr}
Jie Lei, Tamara~L Berg, and Mohit Bansal.
\newblock Detecting moments and highlights in videos via natural language
  queries.
\newblock In M. Ranzato, A. Beygelzimer, Y. Dauphin, P.S. Liang, and J.~Wortman
  Vaughan, editors, {\em Advances in Neural Information Processing Systems},
  volume~34, pages 11846--11858. Curran Associates, Inc., 2021.

\bibitem{Lei2021LessIM}
Jie Lei, Linjie Li, Luowei Zhou, Zhe Gan, Tamara~L. Berg, Mohit Bansal, and
  Jingjing Liu.
\newblock Less is more: Clipbert for video-and-language learning via sparse
  sampling.
\newblock {\em 2021 IEEE/CVF Conference on Computer Vision and Pattern
  Recognition (CVPR)}, pages 7327--7337, 2021.

\bibitem{Li_Guo_Wang_2021}
Kun Li, Dan Guo, and Meng Wang.
\newblock Proposal-free video grounding with contextual pyramid network.
\newblock {\em Proceedings of the AAAI Conference on Artificial Intelligence},
  35(3):1902--1910, May 2021.

\bibitem{Li2020HeroHE}
Linjie Li, Yen-Chun Chen, Yu Cheng, Zhe Gan, Licheng Yu, and Jingjing Liu.
\newblock Hero: Hierarchical encoder for video+language omni-representation
  pre-training.
\newblock {\em ArXiv}, abs/2005.00200, 2020.

\bibitem{li2021learnable}
Yang Li, Si Si, Gang Li, Cho-Jui Hsieh, and Samy Bengio.
\newblock Learnable fourier features for multi-dimensional spatial positional
  encoding.
\newblock In A. Beygelzimer, Y. Dauphin, P. Liang, and J.~Wortman Vaughan,
  editors, {\em Advances in Neural Information Processing Systems}, 2021.

\bibitem{li2022efficientformer}
Yanyu Li, Geng Yuan, Yang Wen, Eric Hu, Georgios Evangelidis, Sergey Tulyakov,
  Yanzhi Wang, and Jian Ren.
\newblock Efficientformer: Vision transformers at mobilenet speed.
\newblock {\em arXiv preprint arXiv:2206.01191}, 2022.

\bibitem{egovlp}
Kevin~Qinghong Lin, Jinpeng Wang, Mattia Soldan, Michael Wray, Rui Yan, Eric~Z
  XU, Difei Gao, Rong-Cheng Tu, Wenzhe Zhao, Weijie Kong, et~al.
\newblock Egocentric video-language pretraining.
\newblock {\em Advances in Neural Information Processing Systems},
  35:7575--7586, 2022.

\bibitem{adamw}
Ilya Loshchilov and Frank Hutter.
\newblock Decoupled weight decay regularization.
\newblock In {\em International Conference on Learning Representations}, 2019.

\bibitem{luo2021clip4clip}
Huaishao Luo, Lei Ji, Ming Zhong, Yang Chen, Wen Lei, Nan Duan, and Tianrui Li.
\newblock Clip4clip: An empirical study of clip for end to end video clip
  retrieval.
\newblock {\em arXiv preprint arXiv:2104.08860}, 2021.

\bibitem{Mun_2020_CVPR}
Jonghwan Mun, Minsu Cho, and Bohyung Han.
\newblock {Local-Global Video-Text Interactions for Temporal Grounding}.
\newblock In {\em Proceedings of the IEEE/CVF Conference on Computer Vision and
  Pattern Recognition (CVPR)}, 2020.

\bibitem{niranjan2020end}
Abhishek Niranjan, Mukesh Sharma, Sai Bharath~Chandra Gutha, and M Shaik.
\newblock End-to-end whisper to natural speech conversion using modified
  transformer network.
\newblock {\em arXiv preprint arXiv:2004.09347}, 2020.

\bibitem{qing2021temporal}
Zhiwu Qing, Haisheng Su, Weihao Gan, Dongliang Wang, Wei Wu, Xiang Wang, Yu
  Qiao, Junjie Yan, Changxin Gao, and Nong Sang.
\newblock Temporal context aggregation network for temporal action proposal
  refinement.
\newblock In {\em Proceedings of the IEEE/CVF Conference on Computer Vision and
  Pattern Recognition}, pages 485--494, 2021.

\bibitem{radford2021learning}
Alec Radford, Jong~Wook Kim, Chris Hallacy, Aditya Ramesh, Gabriel Goh,
  Sandhini Agarwal, Girish Sastry, Amanda Askell, Pamela Mishkin, Jack Clark,
  et~al.
\newblock Learning transferable visual models from natural language
  supervision.
\newblock {\em arXiv preprint arXiv:2103.00020}, 2021.

\bibitem{ramazanova2023owl}
Merey Ramazanova, Victor Escorcia, Fabian~Caba Heilbron, Chen Zhao, and Bernard
  Ghanem.
\newblock Owl (observe, watch, listen): Localizing actions in egocentric video
  via audiovisual temporal context.
\newblock In {\em Proceedings of the IEEE/CVF Conference on Computer Vision and
  Pattern Recognition Workshop (CVPRW)}, 2023.

\bibitem{TACoS_ACL_2013}
Michaela Regneri, Marcus Rohrbach, Dominikus Wetzel, Stefan Thater, Bernt
  Schiele, and Manfred Pinkal.
\newblock {Grounding Action Descriptions in Videos}.
\newblock {\em Transactions of the Association for Computational Linguistics
  (ACL)}, 2013.

\bibitem{Rodriguez_2020_WACV}
{Rodriguez Cristian, Marrese-Taylor Edison, Saleh Fatemeh Sadat, Li Hongdong,
  Gould Stephen}.
\newblock {Proposal-free Temporal Moment Localization of a Natural-Language
  Query in Video using Guided Attention}.
\newblock In {\em Proceedings of the IEEE Winter Conference on Applications of
  Computer Vision (WACV)}, 2020.

\bibitem{Soldan_2022_CVPR}
Mattia Soldan, Alejandro Pardo, Juan~Le\'on Alc\'azar, Fabian Caba, Chen Zhao,
  Silvio Giancola, and Bernard Ghanem.
\newblock Mad: A scalable dataset for language grounding in videos from movie
  audio descriptions.
\newblock In {\em Proceedings of the IEEE/CVF Conference on Computer Vision and
  Pattern Recognition (CVPR)}, pages 5026--5035, June 2022.

\bibitem{soldan2021vlg}
Mattia Soldan, Mengmeng Xu, Sisi Qu, Jesper Tegner, and Bernard Ghanem.
\newblock Vlg-net: Video-language graph matching network for video grounding.
\newblock In {\em Proceedings of the IEEE/CVF International Conference on
  Computer Vision}, pages 3224--3234, 2021.

\bibitem{NIPS2017_attent}
Ashish Vaswani, Noam Shazeer, Niki Parmar, Jakob Uszkoreit, Llion Jones,
  Aidan~N Gomez, \L~ukasz Kaiser, and Illia Polosukhin.
\newblock Attention is all you need.
\newblock In I. Guyon, U.~Von Luxburg, S. Bengio, H. Wallach, R. Fergus, S.
  Vishwanathan, and R. Garnett, editors, {\em Advances in Neural Information
  Processing Systems}, volume~30. Curran Associates, Inc., 2017.

\bibitem{NIPS2017_7181}
Ashish Vaswani, Noam Shazeer, Niki Parmar, Jakob Uszkoreit, Llion Jones,
  Aidan~N Gomez, \L~ukasz Kaiser, and Illia Polosukhin.
\newblock {Attention is All you Need}.
\newblock In {\em Advances in Neural Information Processing Systems (NIPS)},
  2017.

\bibitem{verma2021audio}
Prateek Verma and Jonathan Berger.
\newblock Audio transformers: Transformer architectures for large scale audio
  understanding. adieu convolutions.
\newblock {\em arXiv preprint arXiv:2105.00335}, 2021.

\bibitem{wolf2020transformers}
Thomas Wolf, Lysandre Debut, Victor Sanh, Julien Chaumond, Clement Delangue,
  Anthony Moi, Pierric Cistac, Tim Rault, R{\'e}mi Louf, Morgan Funtowicz,
  et~al.
\newblock Transformers: State-of-the-art natural language processing.
\newblock In {\em Proceedings of the 2020 conference on empirical methods in
  natural language processing: system demonstrations}, pages 38--45, 2020.

\bibitem{xu2021videoclip}
Hu Xu, Gargi Ghosh, Po-Yao Huang, Dmytro Okhonko, Armen Aghajanyan, Florian
  Metze, Luke Zettlemoyer, and Christoph Feichtenhofer.
\newblock Videoclip: Contrastive pre-training for zero-shot video-text
  understanding.
\newblock {\em arXiv preprint arXiv:2109.14084}, 2021.

\bibitem{xu2023boundary}
Mengmeng Xu, Mattia Soldan, Jialin Gao, Shuming Liu, Juan-Manuel
  P{\'e}rez-R{\'u}a, and Bernard Ghanem.
\newblock Boundary-denoising for video activity localization.
\newblock {\em arXiv preprint arXiv:2304.02934}, 2023.

\bibitem{AAAIYangWW0XYH22}
Haosen Yang, Wenhao Wu, Lining Wang, Sheng Jin, Boyang Xia, Hongxun Yao, and
  Hujie Huang.
\newblock Temporal action proposal generation with background constraint.
\newblock In {\em Thirty-Sixth {AAAI} Conference on Artificial Intelligence,
  {AAAI} 2022, Thirty-Fourth Conference on Innovative Applications of
  Artificial Intelligence, {IAAI} 2022, The Twelveth Symposium on Educational
  Advances in Artificial Intelligence, {EAAI} 2022 Virtual Event, February 22 -
  March 1, 2022}, pages 3054--3062. {AAAI} Press, 2022.

\bibitem{Yao2021EfficientDI}
Zhuyu Yao, Jiangbo Ai, Boxun Li, and Chi Zhang.
\newblock Efficient detr: Improving end-to-end object detector with dense
  prior.
\newblock {\em ArXiv}, abs/2104.01318, 2021.

\bibitem{Zeng_2020_CVPR}
Runhao Zeng, Haoming Xu, Wenbing Huang, Peihao Chen, Mingkui Tan, and Chuang
  Gan.
\newblock {Dense Regression Network for Video Grounding}.
\newblock In {\em Proceedings of the IEEE/CVF Conference on Computer Vision and
  Pattern Recognition (CVPR)}, 2020.

\bibitem{vslnet}
Hao Zhang, Aixin Sun, Wei Jing, and Joey~Tianyi Zhou.
\newblock Span-based localizing network for natural language video
  localization.
\newblock In {\em Proceedings of the 58th Annual Meeting of the Association for
  Computational Linguistics}, pages 6543--6554, Online, July 2020. Association
  for Computational Linguistics.

\bibitem{2DTAN_2020_AAAI}
{Zhang Songyang, Peng Houwen, Fu Jianlong, Luo, Jiebo}.
\newblock {Learning 2D Temporal Adjacent Networks for Moment Localization with
  Natural Language}.
\newblock In {\em Proceedings of the AAAI Conference on Artificial
  Intelligence}, 2020.

\bibitem{SSN2017ICCV}
Yue Zhao, Yuanjun Xiong, Limin Wang, Zhirong Wu, Xiaoou Tang, and Dahua Lin.
\newblock Temporal action detection with structured segment networks.
\newblock In {\em ICCV}, 2017.

\end{thebibliography}
}
\end{document}


\title{
Localizing Moments in Long Video Via Multimodal Guidance  \\
(\textit{Supplementary Material})}

{\author{
Wayner Barrios$^{1}$,
\quad Mattia Soldan$^{2}$, 
\quad Alberto Mario Ceballos-Arroyo$^{3}$,\\
\quad Fabian Caba Heilbron$^{4}$,
\quad Bernard Ghanem$^{2}$ \\\\
\small$^{1}$ Dartmouth \quad $^{2}$King Abdullah University of Science and Technology (KAUST)
\small$^{3}$Northeastern University \quad $^{4}$Adobe Research
}}
\maketitle
\ificcvfinal\thispagestyle{empty}\fi

This document provides detailed information on the following topics: \textbf{(A)} Short-form video setup for the baseline models on MAD~\cite{Soldan_2022_CVPR} dataset: Zero-shot CLIP~\cite{Soldan_2022_CVPR}, VLG-Net~\cite{soldan2021vlg}, and Moment-DETR~\cite{moment-detr}. \textbf{(B)} Comprehensive results using Moment-DETR~\cite{moment-detr} and VSL-Net~\cite{vslnet} in Ego4D~\cite{ego4d}. \textbf{(C)} In-depth results of each ablation study presented in the main paper for the Guidance Model. Lastly, \textbf{(D)}  Evaluating the inference time of the Guidance Model.
\begin{table}[!b]{
\centering

\setlength{\tabcolsep}{2.5pt}
\renewcommand{\arraystretch}{1} 
\resizebox{\columnwidth}{!}{%
\begin{tabular}{lllllll}

\toprule
\multirow{2}{*}{\textbf{Model}} & \multicolumn{2}{c}{\textbf{IoU=0.1}} & \multicolumn{2}{c}{\textbf{IoU=0.3}} & \multicolumn{2}{c}{\textbf{IoU=0.5}}\\
\cmidrule{2-7} 
\ &
\textbf{R@1} &
\textbf{R@5} &
\textbf{R@1} &
\textbf{R@5} &
\textbf{R@1} &
\textbf{R@5} \\ \midrule

Zero-shot CLIP~\cite{Soldan_2022_CVPR}       & $37.83$           & $69.53$       & $17.0$1           & $47.63$                                                    & $6.75$            & $25.81$          \\ 
VLG-Net~\cite{soldan2021vlg}                 & $44.70$           & $73.99$        & $23.91$           & $61.81$                                                & $17.59$         & $44.30$          \\ 
Moment-DETR~\cite{moment-detr}              & $\mathbf{46.30}$           & $\mathbf{93.17}$          &     $\mathbf{34.00}$                             & $\mathbf{68.88}$          & $\mathbf{20.12}$          & $\mathbf{44.48}$        \\ \bottomrule
\end{tabular}
}}
\vspace{-.1cm}
\caption{\small\label{tab:30se}{\bf Short-form video setup: $\mathbf{30}$ seconds.} 
The table shows the performance of the three selected grounding baseline models in a short-video setup. In this setting, videos are chunked into $30$-second, non-overlapping windows. Moment-DETR achieves the best performance in all the metrics for this setup, however, it falls behind zero-shot CLIP and VLG-Net in the long-form video setup (see Section $4$ of the main paper).}

\end{table}
\section*{A. Short-form video setup}~\label{short}
This section presents additional results for the short-form moment localization setup. To this end, we use the baseline grounding models reported in the main paper and evaluate them using several short-form video setups, following \cite{Soldan_2022_CVPR}. To be more specific, we first split a single (long) movie into non-overlapping windows (short videos). Then, we assign to each of these windows the ground-truth annotation with the highest temporal overlap. We evaluate five short-form video setups using $30$, $60$, $120$, $180$, and $600$ second windows.
Table \ref{tab:30se} showcases the performance of the three baseline models in the ($30$ seconds) short-form video setup. Moment-DETR~\cite{moment-detr} achieves the highest scores for all metrics. This result indicates that the model is remarkably good for moment localization in short videos, however, it utterly fails to ground moments in video setups with much longer windows (see Table 1 in the main paper).

\begin{figure}[h!]
\centering

\begin{subfigure}[h!]{\linewidth}
\includegraphics[width=\linewidth]{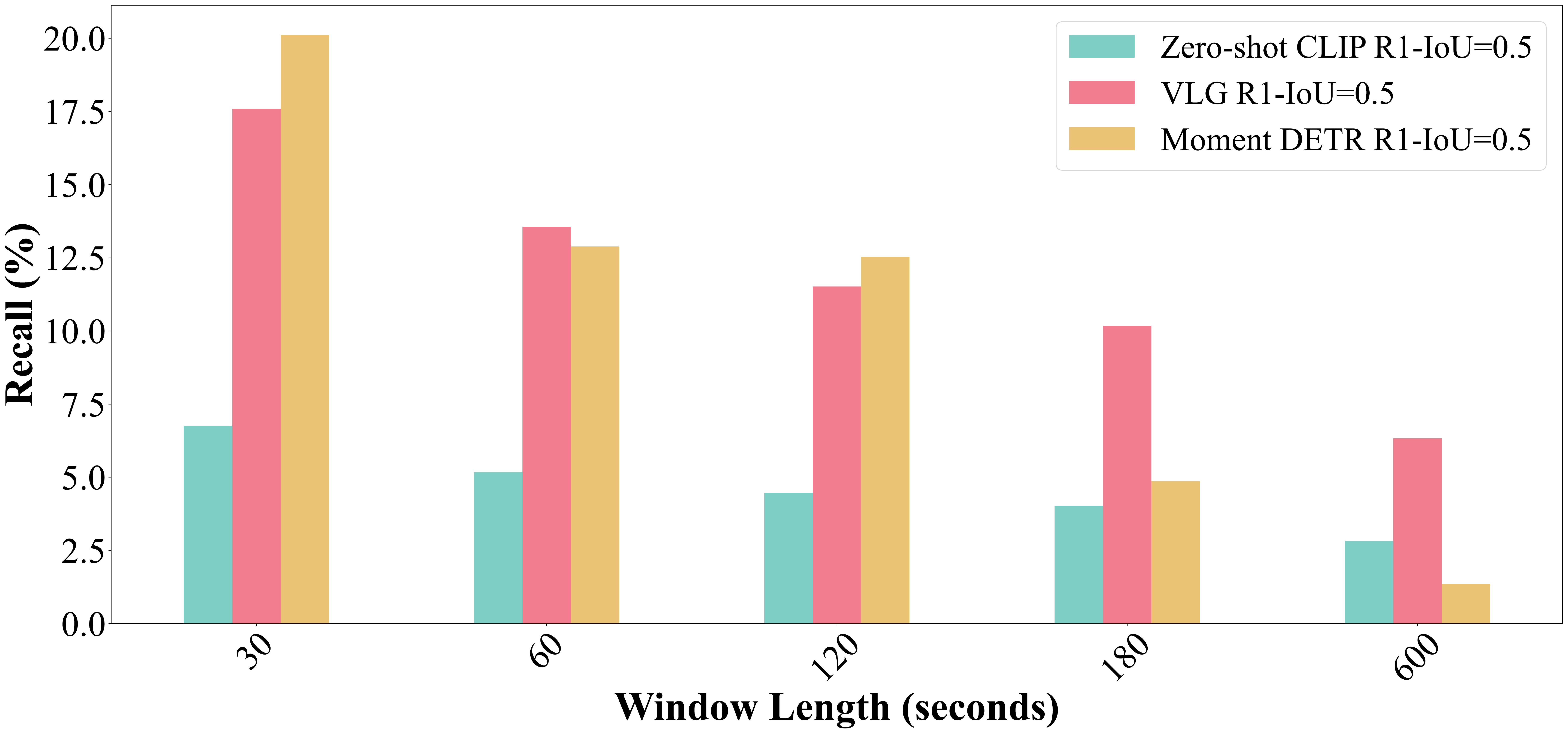}
\small\subcaption{Baseline models performance in short-form video setup at R@$1$-IoU=$0.5$.}
\vspace{0.4cm}
\label{fig:r1_short}
\end{subfigure}
\begin{subfigure}[h!]{\linewidth}
\includegraphics[width=\linewidth]{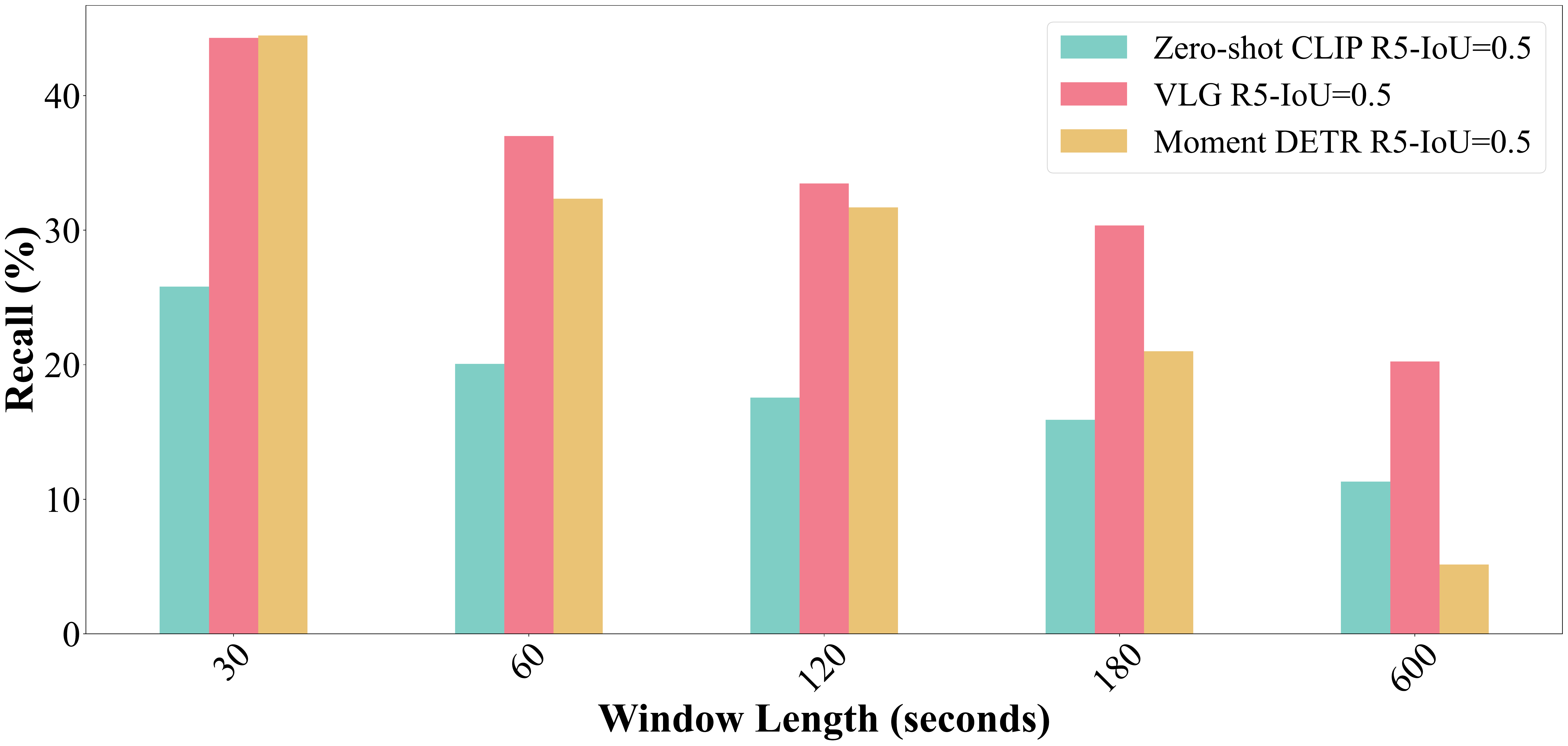}
\small\caption{Baseline models performance in short-form video setup at R@$5$-IoU=$0.5$.}
\label{fig:r5_short}
\end{subfigure}
 \caption{\textbf{Performance trend across different short-form video setups.} Figure (a) illustrates the baseline performance in the short-form video setup at R@$1$ using tIoU=$0.5$. In contrast, Figure (b) displays the performance at R@$5$ with tIoU=$0.5$. The figure also demonstrates how the performance of all baseline models changes as the evaluation window lengthens. Notably, we observe that Moment-DETR is the most affected by long window sizes exceeding $120$ seconds.
 } 
\label{fig:short}
\end{figure}
\begin{table*}[!t]
\centering
\setlength{\tabcolsep}{2.5pt}
\renewcommand{\arraystretch}{1} 
\resizebox{\linewidth}{!}{%
\footnotesize
\begin{tabular}{@{}c@{\hspace{0.4em}} 
l 
c@{\hspace{0.2em}} 
ccccc c@{\hspace{0.2em}} 
ccccc c@{\hspace{0.2em}} 
cccc  c@{}}

\toprule
& \multirow{2}{*}{\textbf{Model}}
& \phantom{} & \multicolumn{5}{c}{\textbf{IoU=0.1}} 
& \phantom{} & \multicolumn{5}{c}{\textbf{IoU=0.3}}
& \phantom{} & \multicolumn{5}{c}{\textbf{IoU=0.5}} \\

\cmidrule{4-8} \cmidrule{10-14} \cmidrule{16-20} 
&
&& \textbf{R@1} & \textbf{R@5} & \textbf{R@10} & \textbf{R@50} & \textbf{R@100}
&& \textbf{R@1} & \textbf{R@5} & \textbf{R@10} & \textbf{R@50} & \textbf{R@100}
&& \textbf{R@1} & \textbf{R@5} & \textbf{R@10} & \textbf{R@50} & \textbf{R@100} \\
\midrule
&Moment-DETR~\cite{moment-detr}  
    && $12.87$ & $37.35$ & $41.32$ & $42.93$ & $-$ 
    && $5.21$ & $17.23$  & $20.73$ & $21.14$ & $-$ 
    && $2.09$ & $4.47$  &  $9.50$ & $9.94$ & $-$  \\
          
&VSL-Net~\cite{vslnet,egovlp}  
    && $16.52$ & $26.59$ & $32.52$ & $47.47$ & $54.16$ 
    && $10.82$ & $18.87$  & $23.21$ & $33.94$ & $40.40$ 
    && $6.81$ & $13.45$  &  $17.09$ & $26.87$ & $31.80$  \\

\cmidrule{1-20}
&$\dagger$Moment-DETR~\cite{moment-detr}  
    && $13.47$ & $\mathbf{39.55}$ & $\mathbf{43.10}$ & $44.63$ & $-$ 
    && $5.83$ & $18.87$  & $21.14$ & $22.41$ & $-$ 
    && $2.53$ & $8.00$  &  $10.02$ & $11.10$ & $-$  \\                                  
&$\dagger$VSL-Net~\cite{vslnet,egovlp}  && 
    $\mathbf{17.11}$ & $31.85$ & $42.13$ & $\mathbf{66.99}$ & $\mathbf{67.20}$ 
    && $\mathbf{11.25}$ & $\mathbf{23.95}$ & $\mathbf{30.25}$ & $\mathbf{53.99}$ & $\mathbf{54.20}$ 
    && $\mathbf{7.20}$ & $\mathbf{16.37}$ & $\mathbf{27.55}$ & $\mathbf{42.80}$ & $\mathbf{43.30}$   \\

\bottomrule
\end{tabular}%
}
\vspace{-.1cm}
\caption{\small\label{tab:ego_grounding}{\bf Benchmarking of grounding methods on the Ego4D dataset.}  
We present four models:Moment-DETR~\cite{moment-detr} and VSL-Net~\cite{vslnet,egovlp} in rows 1 and 2, respectively, without Guidance Model. Rows 3 and 4 demonstrate the impact of our Guidance Model on Moment-DETR's performance and VSL-Net's performance. To denote the utilization of the Guidance Model, we use the symbol $\dagger$. It's important to note that these results were obtained using the best Guidance Model under a query-dependent setup.
}
\end{table*}

Figure \ref{fig:short} presents the performance of the three baselines in the five short-form video grounding setups. Moment-DETR~\cite{moment-detr} has the best performance when grounding moments on $30$ second windows and achieves comparable performance to VLG-Net~\cite{soldan2021vlg} in the other short setups (i.e., $30$, $60$, $120$ seconds). However, the performance gap between Moment-DETR and VLG-Net starts to grow for window lengths greater than $120$ seconds. Furthermore, Moment-DETR is better than zero-shot CLIP in all short-form video setups except for the one with a $600$ second window length, where Moment-DETR exhibits the worst performance. In any case, we must note that all models exhibit a significant drop in performance when attempting to process longer videos.

\textbf{Takeaways:} \textbf{(i)} Moment-DETR works well with short-form videos but fails for videos longer than $120$ seconds, \textbf{(ii)} current state-of-the-art models cannot address the video grounding task in the long-form video setting, and \textbf{(iii)} using the Guidance Model along with Moment-DETR can boost the latter's performance allowing it to compete with state-of-the-art models on the MAD Dataset.
\section*{B. Ego4D Results}
We also evaluated the effectiveness of the Guidance Model on the Ego4D~\cite{ego4d} validation dataset. In Table~\ref{tab:ego_grounding}, we present the performance of Moment-DETR~\cite{moment-detr} and VSL-Net~\cite{vslnet,egovlp} and the boost in performance for the same baseline using the best configuration of our Guidance Model. By reporting Recall@$K$ with IoU=$\theta$, for $K{\in}\{1, 5, 10, 50, 100\}$ and $\theta{\in}\{0.1, 0.3, 0.5\}$, instead of mR@$K$, we are able to provide more detailed results. This allows us to obtain a more fine-grained understanding of the performance of the model and its ability to ground moments accurately by using natural language queries. 

Generally speaking, VSL-Net with the Guidance model performs the best in most of  the metrics. For instance, R@$1$-IoU=$0.5$ increases from $6.81$ to $7.20$, for R@$5$-IoU=$0.3$ increases from $18.87$ to $23.95$ and R@$10$-IoU=$0.1$ increases from $32.52$ to $42.13$. In contrast, Moment-DETR demonstrated notable performance enhancements, illustrated by a significant rise in R@$5$-IoU scores. Specifically, at the IoU threshold of $0.1$, the metric increased from $26.59$ to an impressive $39.55$, while at the more stringent threshold of $0.5$, it improved from $4.47$ to $8.00$. 


\textbf{Takeaway:} The efficacy of the Guidance Model in enhancing the performance of the base grounding model demonstrates its versatility across datasets and its compatibility with a variety of vision-language grounding models.
\begin{table*}[!t]
\resizebox{\linewidth}{!}{%
\centering
\setlength{\tabcolsep}{2.5pt}
\renewcommand{\arraystretch}{1} 
\begin{tabular}{lllllllllllllllll}
\toprule
  \multicolumn{1}{c}{\multirow{2}{*}{\textbf{Model}}}                              &    & \multicolumn{5}{c}{\textbf{IoU=0.1}} & \multicolumn{5}{c}{\textbf{IoU=0.3}} & \multicolumn{5}{c}{\textbf{IoU=0.5}} \\

\cmidrule{3-17} 
 &
\textbf{Query} &
\multicolumn{1}{c}{\textbf{R@1}} &
  \multicolumn{1}{c}{\textbf{R@5}} &
  \multicolumn{1}{c}{\textbf{R@10}} &
  \multicolumn{1}{c}{\textbf{R@50}} &
  \multicolumn{1}{c}{\textbf{R@100}} &
  \multicolumn{1}{c}{\textbf{R@1}} &
  \multicolumn{1}{c}{\textbf{R@5}} &
  \multicolumn{1}{c}{\textbf{R@10}} &
  \multicolumn{1}{c}{\textbf{R@50}} &
  \multicolumn{1}{c}{\textbf{R@100}} &
  \multicolumn{1}{c}{\textbf{R@1}} &
  \multicolumn{1}{c}{\textbf{R@5}} &
  \multicolumn{1}{c}{\textbf{R@10}} &
  \multicolumn{1}{c}{\textbf{R@50}} &
  \multicolumn{1}{c}{\textbf{R@100}}  &
\midrule
\multirow{3}{*}{Zero-shot CLIP~\cite{Soldan_2022_CVPR}} 
& \xmark &
  $6.65$ &
  $14.80$ &
  $19.79$ &
  $36.30$ &
  $45.59$ &
  $3.19$ &
  $9.88$ &
  $13.84$ &
  $27.47$ &
  $35.33$ &
  $1.35$ &
  $5.31$ &
  $8.03$ &
  $18.06$ &
  $23.93$ \\
& AG 
& $6.84$ 
& $15.19$    
& $20.28$    
& $37.88$    
& $\mathbf{45.80}$    
& $3.89$    
& $9.93$    
& $15.22$    
& $27.80$   
& $36.60$   
& $1.41$    
& $5.49$    
& $9.29$   
& $17.44$    
& $24.12$    \\
& DE 
&  $\mathbf{9.05}$  
&  $\mathbf{18.00}$   
&  $\mathbf{22.76}$   
&  $\mathbf{37.24}$  
&  $44.37$   
&  $\mathbf{4.52}$   
&  $\mathbf{12.60}$   
&  $\mathbf{16.83}$   
&  $\mathbf{30.13}$   
&  $\mathbf{37.02}$   
&  $\mathbf{2.01}$   
&  $\mathbf{7.14}$   
&  $\mathbf{10.65}$   
&  $\mathbf{21.75}$  
&  $\mathbf{27.76}$  \\
\midrule
\multirow{3}{*}{VLG-Net~\cite{soldan2021vlg}}        
& \xmark  
& $3.43$    
& $11.32$    
& $17.19$    
& $37.80$    
& $48.63$    
& $2.56$    
& $8.92$    
& $13.92$    
& $32.38$    
& $42.34$    
& $1.49$    
& $5.54$    
& $9.01$    
& $22.93$    
& $31.34$ \\
& AG 
& $3.89$    
& $14.31$   
& $18.69$    
& $39.37$    
& $49.87$    
& $2.82$   
& $9.30$    
& $15.24$   
& $35.22$   
& $44.19$   
& $1.70$   
& $7.12$   
& $9.87$  
& $24.00$   
& $34.16$    \\
& DE 
& $\mathbf{5.37}$    
& $\mathbf{15.52}$    
& $\mathbf{22.37}$    
& $\mathbf{42.58}$    
& $\mathbf{52.72}$    
& $\mathbf{4.01}$    
& $\mathbf{12.56}$    
& $\mathbf{18.58}$    
& $\mathbf{37.33}$    
& $\mathbf{46.44}$    
& $\mathbf{2.31}$    
& $\mathbf{8.16}$    
& $\mathbf{12.57}$    
& $\mathbf{28.14}$    
& $\mathbf{36.46}$   \\
\midrule
\multirow{3}{*}{Moment-DETR~\cite{moment-detr}}    
& \xmark  
& $0.28$    
& $1.44$    
& $2.62$    
& $10.41$    
& $18.21$    
& $0.20$    
& $1.07$    
& $1.87$   
& $7.61$    
& $13.18$    
& $0.12$    
& $0.62$    
& $1.05$    
& $4.47$    
& $7.93$    \\
& AG 
& $0.65$  
& $2.11$    
& $4.54$   
& $18.44$    
& $32.34$   
& $0.59$   
& $1.78$   
& $2.83$   
& $14.22$   
& $22.97$   
& $0.36$   
& $1.02$   
& $1.45$   
& $8.32$   
& $12.54$   \\
& DE 
&  $\mathbf{4.84}$   
&  $\mathbf{15.34}$  
&  $\mathbf{22.80}$   
&  $\mathbf{46.01}$  
&  $\mathbf{57.06}$   
&  $\mathbf{3.69}$  
&  $\mathbf{11.95}$   
&  $\mathbf{17.99}$   
&  $\mathbf{37.10}$   
&  $\mathbf{46.19}$  
&  $\mathbf{2.17}$ 
&  $\mathbf{7.25}$ 
&  $\mathbf{10.93}$  
&  $\mathbf{22.62}$  
&  $\mathbf{28.09}$  \\
\bottomrule
\end{tabular}
}
\caption{\label{tab:agnosti}{\bf Describable windows (full metrics).} In this table, we report Recall@$K$ on the validation partition of MAD for the baseline models under three settings: without any guidance (rows 1, 4, and 7), with query agnostic guidance (AG), and with query dependent guidance (DE). All three baselines benefit the most from using query-dependent guidance, and modest performance improvement by query-agnostic setup. Our findings suggest that the query-dependent approach yields superior performance despite its high computational cost. However, adopting a query-agnostic setup can also enhance performance at a lower computational cost. }

\vspace{-.3cm}

\end{table*}
\begin{table}[b]

\centering
\setlength{\tabcolsep}{2.5pt}
\renewcommand{\arraystretch}{1} 
\resizebox{\columnwidth}{!}{%
\footnotesize
\begin{tabular}{@{}c@{\hspace{0.4em}} 
l 
c@{\hspace{0.2em}} 
ccccc c@{\hspace{0.2em}} 
ccccc c@{\hspace{0.2em}} 
cccc  c@{}}

\toprule
& \multirow{2}{*}{\textbf{Model}}
& \phantom{} & \multicolumn{3}{c}{\textbf{IoU=0.1}} 
& \phantom{} & \multicolumn{3}{c}{\textbf{IoU=0.3}}
& \phantom{} & \multicolumn{3}{c}{\textbf{IoU=0.5}} \\

\cmidrule{4-6} \cmidrule{8-10} \cmidrule{12-14} 
&
&& \textbf{R@1} & \textbf{R@5} & \textbf{R@10} 
&& \textbf{R@1} & \textbf{R@5} & \textbf{R@10} 
&& \textbf{R@1} & \textbf{R@5} & \textbf{R@10}  \\
\midrule
          
&Zero-shot CLIP~\cite{Soldan_2022_CVPR}  
    && $5.43$ & $11.87$ & $15.98$ 
    && $2.69$ & $7.88$  & $11.06$ 
    && $1.15$ & $4.54$  &  $6.86$  \\
&VLG-Net~\cite{soldan2021vlg}  
    && $2.15$ & $7.65$ & $12.00$ 
    && $1.50$ & $5.84$  & $9.56$ 
    && $0.95$ & $3.83$  &  $6.43$  \\
&Moment-DETR~\cite{moment-detr} 
    && $0.24$ & $0.97$ & $1.78$ 
    && $0.13$ & $0.76$  & $0.85$ 
    && $0.08$ & $0.21$  &  $0.09$  \\

\midrule                            
&$\dagger$Zero-shot CLIP~\cite{Soldan_2022_CVPR}
    && $\mathbf{7.39}$ & $\mathbf{14.90}$ & $\mathbf{19.09}$ 
    && $\mathbf{3.76}$ & $\mathbf{10.32}$  & $\mathbf{13.95}$ 
    && $1.77$ & $5.93$  &  $9.00$  \\
&$\dagger$VLG-Net~\cite{soldan2021vlg}
    && $3.31$ & $10.17$ & $16.06$
    && $2.21$ & $7.83$  & $13.09$ 
    && $1.45$ & $\mathbf{6.20}$  & $\mathbf{10.24}$ \\
&$\dagger$Moment-DETR~\cite{moment-detr} 
    && $3.71$ & $11.37$ & $17.94$
    && $2.84$ & $8.53$  & $13.62$ 
    && $\mathbf{2.04}$ & $5.41$  & $8.98$ \\
\bottomrule
\end{tabular}%
}
\vspace{.2cm}
\caption{\small\label{tab:actionless}{\bf Describable windows beyond actions.} We compare the performance of our baseline models with the guidance module ($\dagger$) and without guidance on actionless queries only. For this purpose, we extract actionless queries from MAD~\cite{Soldan_2022_CVPR} test set and evaluate using mR@$K$ for $K{\in}\{1, 5, 10\}$. The results showcase that  our guidance method is not solely learning action-based concepts.  }
\end{table}

\begin{table*}[!t]
\centering
\setlength{\tabcolsep}{2.5pt}
\renewcommand{\arraystretch}{1} 
\centering
\resizebox{\linewidth}{!}{%
\footnotesize
\begin{tabular}{llllllllllllllllll}
\toprule
\multirow{2}{*}{\textbf{Model}}       & \multicolumn{2}{l}{\textbf{Modalities}} & \multicolumn{5}{c}{\textbf{IoU=0.1}} & \multicolumn{5}{c}{\textbf{IoU=0.3}} & \multicolumn{5}{c}{\textbf{IoU=0.5}} \\
\cmidrule{4-18} 
 &
  \textbf{Audio} &
  \textbf{Visual} &
  \multicolumn{1}{c}{\textbf{R@1}} &
  \multicolumn{1}{c}{\textbf{R@5}} &
  \multicolumn{1}{c}{\textbf{R@10}} &
  \multicolumn{1}{c}{\textbf{R@50}} &
  \multicolumn{1}{c}{\textbf{R@100}} &
  \multicolumn{1}{c}{\textbf{R@1}} &
  \multicolumn{1}{c}{\textbf{R@5}} &
  \multicolumn{1}{c}{\textbf{R@10}} &
  \multicolumn{1}{c}{\textbf{R@50}} &
  \multicolumn{1}{c}{\textbf{R@100}} &
  \multicolumn{1}{c}{\textbf{R@1}} &
  \multicolumn{1}{c}{\textbf{R@5}} &
  \multicolumn{1}{c}{\textbf{R@10}} &
  \multicolumn{1}{c}{\textbf{R@50}} &
  \multicolumn{1}{c}{\textbf{R@100}} &
  \midrule
\multirow{4}{*}{Zero-shot CLIP~\cite{Soldan_2022_CVPR}} &
  \xmark &
  \xmark  &
  $6.65$ &
  $14.80$ &
  $19.79$ &
  $36.30$ &
  $\mathbf{45.59}$ &
  $3.19$ &
  $9.88$ &
  $13.84$ &
  $27.47$ &
  $35.33$ &
  $1.35$ &
  $5.31$ &
  $8.03$ &
  $18.06$ &
  $23.93$ \\
& \cmark           
& \xmark             
&$6.75$  
&$16.08$
& $21.67$   
& $36.51$   
& $45.40$   
& $3.50$   
& $10.85$   
& $15.72$   
& $28.45$  
& $35.49$
& $1.63$   
& $6.23$   
& $9.28$   
& $19.31$  
& $25.07$  \\
& \xmark               
& \cmark            
& $8.33$  
& $17.42$   
& $22.16$   
& $36.91$  
& $44.10$   
& $4.21$  
& $11.93$   
& $16.16$  
& $29.51$  
& $36.47$   
& $1.89$   
& $6.69$   
& $10.01$  
& $21.23$   
& $27.32$  \\
& \cmark             
& \cmark          
&  $\mathbf{9.05}$  
&  $\mathbf{18.00}$   
&  $\mathbf{22.76}$   
&  $\mathbf{37.24}$  
&  $44.37$   
&  $\mathbf{4.52}$   
&  $\mathbf{12.60}$   
&  $\mathbf{16.83}$   
&  $\mathbf{30.13}$   
&  $\mathbf{37.02}$   
&  $\mathbf{2.01}$   
&  $\mathbf{7.14}$   
&  $\mathbf{10.65}$   
&  $\mathbf{21.75}$  
&  $\mathbf{27.76}$  \\
\midrule
\multirow{4}{*}{VLG-Net~\cite{soldan2021vlg}}     
& \xmark              
& \xmark             
& $3.43$    
& $11.32$    
& $17.19$    
& $37.80$    
& $48.63$    
& $2.56$    
& $8.92$    
& $13.92$    
& $32.38$    
& $42.34$    
& $1.49$    
& $5.54$    
& $9.01$    
& $22.93$    
& $31.34$ \\
& \cmark             
& \xmark            
& $4.43$    
& $13.21$   
& $19.20$    
& $38.70$    
& $48.93$    
& $3.34$    
& $10.67$   
& $15.80$    
& $33.20$    
& $42.18$    
& $2.03$    
& $ 6.81$    
& $10.28$    
& $23.96$    
& $31.70$    \\
& \xmark              
& \cmark            
& $4.85$    
& $14.56$   
& $21.15$    
& $41.73$   
& $51.70$    
& $3.60$    
& $11.55$    
& $17.45$    
& $36.43$    
& $45.55$    
& $2.04$    
& $7.30$    
& $11.62$    
& $26.86$    
& $35.18$    \\
& \cmark            
& \cmark            
& $\mathbf{5.37}$    
& $\mathbf{15.52}$    
& $\mathbf{22.37}$    
& $\mathbf{42.58}$    
& $\mathbf{52.72}$    
& $\mathbf{4.01}$    
& $\mathbf{12.56}$    
& $\mathbf{18.58}$    
& $\mathbf{37.33}$    
& $\mathbf{46.44}$    
& $\mathbf{2.31}$    
& $\mathbf{8.16}$    
& $\mathbf{12.57}$    
& $\mathbf{28.14}$    
& $\mathbf{36.46}$   \\
\midrule
\multirow{4}{*}{Moment-DETR~\cite{moment-detr}} 
& \xmark              
& \xmark             
& $0.28$    
& $1.44$    
& $2.62$    
& $10.41$    
& $18.21$    
& $0.20$    
& $1.07$    
& $1.87$   
& $7.61$    
& $13.18$    
& $0.12$    
& $0.62$    
& $1.05$    
& $4.47$    
& $7.93$    \\
& \cmark             
& \xmark             
& $1.41$    
& $5.72$    
& $9.67$    
& $27.59$   
& $39.10$    
& $1.05$    
& $4.31$    
& $7.32$    
& $21.66$    
& $31.02$    
& $0.62$    
& $2.56$   
& $4.35$    
& $12.91$    
& $18.49$   \\
& \xmark             
& \cmark            
& $4.24$    
& $14.41$    
& $21.74$    
& $44.77$    
& $56.11$    
& $3.13$    
& $11.07$    
& $16.84$   
& $35.99$   
& $45.26$    
& $1.86$
& $6.58$    
& $9.98$    
& $21.64$    
& $27.28$   \\

& \cmark             
& \cmark           
&  $\mathbf{4.84}$   
&  $\mathbf{15.34}$  
&  $\mathbf{22.80}$   
&  $\mathbf{46.01}$  
&  $\mathbf{57.06}$   
&  $\mathbf{3.69}$  
&  $\mathbf{11.95}$   
&  $\mathbf{17.99}$   
&  $\mathbf{37.10}$   
&  $\mathbf{46.19}$  
&  $\mathbf{2.17}$ 
&  $\mathbf{7.25}$ 
&  $\mathbf{10.93}$  
&  $\mathbf{22.62}$  
&  $\mathbf{28.09}$  \\
\bottomrule
\end{tabular}
}

\caption{\small\label{tab:modalities}{\bf Modality comparison for the Guidance Model (full metrics)}. This table contains an ablation study on the modalities to be used in the Guidance Model. In the first row of each box (rows 1, 5, and 9), we report the baseline performances without score fusion. The Guidance Model uses three different combinations of modalities: (i) audio and text; (ii) video and text; and (iii) audio, video, and text. We evaluate all model configurations on the validation partition of the MAD dataset, using Recall@$K$ with IoU=$\theta$ for $K{\in}\{1, 5, 10, 50, 100\}$ and $\theta{\in}\{0.1, 0.3, 0.5\}$. Generally speaking, the Guidance Model can boost the baseline results under all configurations, and it achieves the best overall performance when combining all modalities.}

\end{table*}
\section*{C. Ablation Study}
We focused our ablation studies on two key factors: \textbf{(i)} selecting the modalities (visual and/or audio), \textbf{(ii)} comparing query-agnostic versus query-dependent guidance performance using the MAD~\cite{Soldan_2022_CVPR} dataset. Additionally, we extensively investigated actionless moments, which distinguishes our implementation from the temporal proposals method~\cite{SST,SSN2017ICCV,AAAIYangWW0XYH22,HAN2022217,qing2021temporal}.
\subsection*{C.1. Modality Fusion in Guidance Model}

As shown in Table \ref{tab:modalities}, the Guidance Model can improve the baseline results, especially when combining all modalities (audio, video, and text). Instead of using mR@$K$ as reported in Table 3 of the main paper, here we report Recall@$K$ with IoU=$\theta$, for $K{\in}\{1, 5, 10, 50, 100\}$ and $\theta{\in}\{0.1, 0.3, 0.5\}$, enabling us to carry out a more fine-grained analysis. For example, when leveraging all three modalities, the performance of Zero-shot CLIP grows from $6.65\%$ to $9.05\%$ for R@$1$-IoU=$0.1$ and $5.31\%$ to $7.14\%$ for R@$5$-IoU=$0.5$. VLG-Net~\cite{soldan2021vlg} also achieves better performance when using either two or three modalities. For example, R@$1$-IoU=$0.3$ grows from $2.56\%$ to $4.01\%$ under the best configuration. On the other hand, the grounding model that is the most benefited by the Guidance Model is Moment-DETR~\cite{moment-detr}. As discussed in the previous section, this model performs poorly on the grounding task with long-form videos; however, by combining it with the Guidance Model, it can reach a performance that is competitive with state-of-the-art approaches. For instance, Moment-DETR~\cite{moment-detr} beats zero-shot CLIP~\cite{Soldan_2022_CVPR} in the R@$1$-IoU=$0.5$ metric for the best configuration: the former achieves $2.17\%$, while the latter gets $2.01\%$. Moreover, the R@$K$ gap between the two models grows significantly with values of $K$ greater than $10$.

\textbf{Takeaway:} We observe that using audio alone, without visual input, results in only modest performance improvements for Zero-shot CLIP, VLG-Net, and Moment-DETR. On the other hand, incorporating visual cues can enhance the performance of all baselines. Nevertheless, the best overall performance is achieved by combining audio, visual and text cues.

\subsection*{C.2. Describable Windows}

Although the first ablation study suggests that combining audio and video inputs as a representation is powerful, the current Guidance Model used is query-dependent and relies on textual queries to identify irrelevant windows. Our method could be made more efficient by identifying windows that are non-describable regardless of the input query, allowing the Guidance Model to process the video/audio streams only once. To investigate an efficient approach, we devise a query-agnostic Guidance Model that does not process any textual query as part of its input.

In Table \ref{tab:agnosti}, we report the full results from which we derived Table 4 of the main paper. Using audio-visual cues alone (that is, a query-agnostic setup) already leads to improvements for VLG-Net~\cite{soldan2021vlg} and Moment-DETR~\cite{moment-detr}. On the other hand, employing query-dependent guidance plays a vital role in boosting the performance of all three baseline models, as it helps reduce the search space for the video grounding task. For instance, VLG-Net~\cite{soldan2021vlg} goes from $11.32\%$ to $14.31\%$ {R@$5$-IoU=$0.1$} when employing query agnostic guidance, and from $11.32\%$ to $15.52\%$ R@$5$-IoU=$0.1$ when employing query dependent guidance. Similarly, the performance of Moment-DETR~\cite{moment-detr} for most metrics also grows: it goes from $2.62\%$ to $4.54\%$ R@$10$-IoU=$0.1$ with query agnostic guidance, and from $2.62\%$ to $22.80\%$ with query dependent guidance. Lastly, we observe that zero-shot CLIP only displays a noticeable performance boost when leveraging query-dependent guidance. For example, R@$5$-IoU=$0.3$ increases from $9.88\%$ to $12.60\%$ in the query-dependent setting, and from $9.88\%$ to  $9.93\%$ in the query-agnostic setting.

\textbf{Takeaway:} Our findings suggest that the query-dependent setup delivers better performance, but at the expense of increased computational cost as the number of queries increases. Conversely, the query-agnostic setup is computationally efficient since the Guidance Model processes the video and audio only once, making it ideal for real-time or low-resource scenarios.

\subsection*{C.3. Describable Windows beyond actions.}

We proposed a new evaluation setup that highlights the differences between our guidance model and proposal models. 
Specifically, we consider MAD~\cite{Soldan_2022_CVPR} queries that do not contain verbs and therefore do not refer to actions. We acknowledge that interpreting this type of query is a challenging task, distinguishing our approach from other computer vision methods like temporal proposals~\cite{SST,SSN2017ICCV,AAAIYangWW0XYH22,HAN2022217,qing2021temporal}, which do not deal with moments lacking any action element. Thus, to confirm our hypothesis, we present in Table~\ref{tab:actionless} the performance improvement that our Guidance Model offers for actionless queries. The Guidance Model demonstrated a boost in performance in all of the baselines. For example, R@$1$-IoU$0.1$ for Zero-shot CLIP improved from $5.43\%$ to $7.39\%$, VLG-Net at R@$10$-IoU$0.3$ improved from $9.56\%$ to $13.09\%$, and Moment-DETR from $0.21\%$ to $5.41\%$ in R@$5$-IoU$0.5$. 

\textbf{Takeaway:} The higher performance of the Guidance Model for queries without actions is clear evidence that it is not solely focused on actions, but can also handle queries that involve describing the environment and its characteristics (adjectives, nouns). The above statement sets our approach apart from proposal-based methods, since proposals methods~\cite{SST,SSN2017ICCV,AAAIYangWW0XYH22,HAN2022217,qing2021temporal} focus only on actions. 

\begin{table}[!t]
\centering
\setlength{\tabcolsep}{2.5pt}
\renewcommand{\arraystretch}{1} 
\resizebox{\columnwidth}{!}{%
\footnotesize
\vspace{-.1cm}
\begin{tabular}{@{}lccccc@{}}
\toprule 

  \multirow{2}{*}{\textbf{Method} }           
& \textbf{Inference}
& \textbf{Avg. Temp.} 
& \multirow{2}{*}{\textbf{ \# Queries$\times$movies}} 
& \textbf{Number of }\\  

& \textbf{Time}
& \textbf{Windows per movie}
& 
& \textbf{Params (M)} \\

\midrule

\textbf{Grounding}: VLG-Net 
& $18$ Hours   
& $545$   
& $72044\times112$  
& $7.5$ \\


\textbf{Grounding}: Moment-DETR               
& $4.2$ Hours  
& $545$      
& $72044\times112$     
& $4.8$ \\

\midrule

\textbf{Guidance}: Query-agnostic  
& ${\sim}2$ Minutes
& $1091$    
& $- \times 112$       
& $4.8$ \\

\textbf{Guidance}: Query-dependent 
& $8.3$ Hours  
& $1091$     
& $72044\times112$     
& $5.0$ \\

\bottomrule
\end{tabular}%
}
\vspace{.001cm}
\caption{\footnotesize{\label{tab:reb1}{\bf Inference time on the MAD~\cite{Soldan_2022_CVPR} dataset.} Query-agnostic needs a forward pass per movie, while dependent is proportional to the no. of queries/temporal windows. We don't include CLIP~\cite{radford2021learning,Soldan_2022_CVPR} as our approach leverages the model's own pre-computed features making the comparison unfair. }}
\vspace{-1.05cm}
\end{table}








\section*{D. Inference Time.}


Table \ref{tab:reb1} displays inference time for the baseline video grounding models and the Guidance Models. The query-agnostic configuration is intended for situations with restricted computational capabilities. It only requires one sliding-window forward pass per movie, covering 112 movies in total. On the other hand, the query-dependent configuration requires more computational power, as the number of forward passes is directly linked to the total number of queries (72,000) and temporal windows.

A potential direction for future exploration in this paper involves improving inference time efficiency by strategically employing the Guidance Model to pre-filter windows before grounding. This concept revolves around the careful prioritization of recall or precision, depending on the specific needs of the given application. By using the abilities of the Guidance Model, we can choose to handle windows that are most important, thereby reducing the total time for inference while maintaining desired performance standards. This potential also offers an opportunity to tailor the processing pipeline for different applications' unique traits, finding a balanced compromise between computational efficiency and outcome accuracy.


{\small
\bibliographystyle{ieee_fullname}
\bibliography{egbib}
}